\documentclass[12pt]{article}

\usepackage{amsthm,enumerate}
\usepackage{mathtools}
\usepackage[normalem]{ulem}
\usepackage[T1]{fontenc}
\usepackage[utf8]{inputenc}
\usepackage[thinc]{esdiff}
\usepackage{color}
\usepackage{fancyhdr}
\usepackage{calc}
\usepackage{url}
\usepackage{xcolor}
\usepackage{xspace}

\usepackage[top=80pt,bottom=85pt,left=85pt,right=85pt]{geometry}
\usepackage{amsmath,amssymb,graphicx,float,color,tikz,cite,savesym,wasysym,wrapfig}
\usepackage{caption}
\usepackage[debug,pageanchor=false]{hyperref}
\definecolor{link}{rgb}{1,0.45,0.05}
\hypersetup{colorlinks=true,linkcolor=link,citecolor=link,urlcolor=link,linktocpage}

\usepackage{bm}
\usepackage{amssymb}

\usepackage{subfig}

\usepackage{verbatim}
\usepackage{algorithm}
\usepackage{algorithmic}

\usepackage{amssymb,amsmath,amsthm}

\newtheorem{theorem}{Theorem}

\usepackage[utf8]{inputenc} 
\usepackage[T1]{fontenc}    
\usepackage{hyperref}       
\usepackage{url}            
\usepackage{booktabs}       
\usepackage{amsfonts}       
\usepackage{nicefrac}       
\usepackage{microtype}      
\usepackage{xcolor}         
\usepackage{graphicx}
\usepackage{amsmath}
\usepackage{subfig}

\title{\bf Improving Energy Conserving Descent for Machine Learning:  Theory and Practice}

\author{
  G.~Bruno De Luca$^{* 1}$\\
  \texttt{\small gbdeluca@stanford.edu} \\
  \and
   Alice Gatti$^{*}$ \\
   \texttt{\small agatti0@proton.me} \\
   \and
   Eva Silverstein$^{* 1}$ \\
  \texttt{\small evas@stanford.edu}
}
\date{\small $^1$Stanford Institute for Theoretical Physics, Stanford University, Stanford, CA 94306, USA\\
\footnote{ Equal contribution.}}

\begin{document}

\maketitle

\begin{abstract}
\noindent We develop the theory of Energy Conserving Descent (ECD) and introduce ECDSep, a gradient-based optimization algorithm able to tackle convex and non-convex optimization problems. The method is based on the novel ECD framework of optimization as physical evolution of a suitable chaotic energy-conserving dynamical system, enabling analytic control of the distribution of results -- dominated at low loss -- even for generic high-dimensional problems with no symmetries.  Compared to previous realizations of this idea, we exploit the theoretical control to improve both the dynamics and chaos-inducing elements, enhancing performance while simplifying the hyper-parameter tuning of the optimization algorithm targeted to different classes of problems.    
We empirically compare with popular optimization methods such as SGD, Adam and AdamW on a wide range of machine learning problems, finding competitive or improved performance compared to the best among them on each task.  We identify limitations in our analysis pointing to possibilities for additional improvements.

\end{abstract}

\section{Introduction}
The novel {\it Energy Conserving Descent} (ECD) framework for optimization introduced in \cite{BBI}\ was shown in small experiments to be competitive with standard algorithms such as stochastic gradient descent with momentum (SGDM) while offering a theoretically predictable distribution of optimization results.  {\bf Our contribution} is to leverage and improve this theoretical understanding to systematically enhance the performance and demonstrate an overall advantage compared to standard methods (Adam \cite{kingma2017adam}, AdamW \cite{adamw_paper} and SGDM \cite{sgd_paper}) on a diverse suite of small/medium-scale machine learning benchmarks.  This succeeds via a new parameter enabling stronger concentration of the results at small loss, combined with the use of a separable Hamiltonian as in previously developed sampling versions of the algorithm \cite{ESH,Robnik:2022bzs}, which allow a more effective integration.
Overall in our experiments, ECD is competitive  with the better of Adam and SGDM, without need for learning rate (lr) scheduling required by SGDM for competitive performance.
Along the way we  describe limitations and derive theoretical predictions for further improvement.  

\section{ECD Theory and formulaic improvements}\label{sec:theory-formulaic}

ECD is formulated, in analogy to classical mechanics in physics, as a discretization of chaotic energy-conserving Hamiltonian evolution on a $2n$-dimensional phase space of positions $\bf{\Theta}$ (e.g. an $n$-dimensional space of neural network weights and biases) and momenta $\bf{\Pi}$.   The distribution of results in $\bf{\Theta}$ --  either along a given trajectory, or among multiple trajectories -- is given by 

\begin{equation}\label{eq:measure-general}
  \mu({\bf{\Theta}})=  \int d^n\Pi \delta(H({\bf{\Pi}},{\bf{\Theta}})-E)
  = \frac{\Omega_{n-1} |\Pi|^{n-1}}{| \partial_{\Pi} H |} = \frac{\Omega_{n-1} |\Pi|^{n-1}}{| d{\bf{\Theta}}/dt|},
\end{equation}
where $E$ is a constant, the conserved energy, and $H(\bf{\Pi},\bf{\Theta})$ is a time-independent Hamiltonian function which depends on the objective function $F(\bf{\Theta})$ and on the magnitude $|\Pi|$ of the momentum.  The continuum evolution equations are $d{\bf\Pi}/dt = -\partial H/\partial{\bf\Theta}, d{\bf{\Theta}}/dt = \partial H/\partial {\bf{\Pi}}$.  An example we will focus on is a Hamiltonian describing the kinetic energy of a particle with ${\bf{\Theta}}$-dependent mass $\sim 1/V[F({\bf{\Theta}})]$ \cite{ESH, BBI, Robnik:2022bzs} 
\begin{align}
    H({\bf{\Pi}},{\bf{\Theta}}) &= V[F({\bf{\Theta}})] \Pi^2 = \frac{1}{V} {\dot{\bf{\Theta}}}^2 \label{eq:RLeta} \\
V[F(\bf{\Theta})] & \propto (F(\mathbf{\Theta}) - F_0)^\eta, ~~~~ \eta\geqslant 1 \label{eq:F-F0-eta}
\end{align}
along with energy-conserving momentum rotations to enhance chaotic behavior \cite{BBI} (see \S\ref{sub:bounces}).
The continuum-equivalent dynamics obtained from the logarithm of (\ref{eq:RLeta}) separates the position and momentum dependencies allowing more robust numerical integration \cite{ESH, Robnik:2022bzs}.   For (\ref{eq:RLeta}), the measure (\ref{eq:measure-general}) gives
\begin{equation}\label{eq:measure-RLeta}
    \mu({\bf{\Theta}}) = E^{(n-2)/2} \frac{\pi^{n/2}}{\Gamma(n/2)} V[F({\bf{\Theta}})]^{-n/2} .
\end{equation}
We would like to concentrate this measure in desirable regions of the objective.  From (\ref{eq:measure-RLeta}) and (\ref{eq:F-F0-eta}) we see that increasing $\eta$ above 1 (the value taken in \cite{BBI}) accentuates the concentration of measure as close as possible to $F\simeq F_0$, suggesting improved performance with increasing $\eta$.
However, we also must ensure that the enhanced measure -- which corresponds to smaller velocity -- does not come at the cost of excessive slowdown of the motion on the way to the objective.  It is useful to analyze this in concert with the energy conservation equation which determines the speed in terms of the objective (using (\ref{eq:RLeta})-(\ref{eq:F-F0-eta}))  
\begin{equation}\label{eq:E-cons-speed-thetadot}
    {\dot{\bf \Theta}}^2 = E V = E (F(\mathbf{\Theta})-F_0)^\eta.
\end{equation}
To avoid exponential suppression of the speed at large $\eta$, this motivates considering also a regime of $F_0$ values for which $|F-F_0|\gtrsim 1$ at large $\eta$.  { This depends on the minimal value of the objective, $F_{min}$. This is a priori unknown, but in simple cases is well approximated by 0; more generally it can be found via an adaptive procedure \cite{BBI}. }  
Studying the quantity (\ref{eq:measure-RLeta}) in the minimal quadratic basin of the loss $F \sim F^{(2)}_{ij} \Theta^i\Theta^j${$+ F_{min}$} yields  
\begin{equation}\label{eq:measureeta}
    \int d^n\Theta \mu({\bf{\Theta}}) \sim \int |{\bf{\Theta}}|^{n-1} d|{\bf{\Theta}}|(F^{(2)}_{ij} \Theta^i\Theta^j{+ F_{min}}-F_0)^{-\eta n/2}.
\end{equation}
To get a sense of the dominant contributions to the measure, taking into account the geometrical $|{\bf \Theta}|^{n-1}$ factor, let us consider for simplicity an isotropic basin with $F_{ij}=F_2 \delta_{ij}$.  The integral in (\ref{eq:measureeta}) has a saddle point (a peak of the integrand) at 
{\begin{equation}\label{eq:theta-star}
    {\bf \Theta_*}^2=\frac{(F_{min}-F_0)(n-1)}{F_2 (1+n (\eta-1))}.
\end{equation}}

This is where the measure is concentrated (somewhat analogous to the `typical set' in Hamiltonian sampling \cite{2017arXiv170102434B, Robnik:2022bzs}), and it indeed approaches the bottom of the basin for large $\eta$.
Consistently with this, the speed (\ref{eq:E-cons-speed-thetadot}) remains finite and nonzero generically

\begin{equation}\label{eq:speedstar}
    |\dot{{\bf\Theta}}|_{{\bf\Theta}_*} =\sqrt{E}\left(\frac{(F_{min}-F_0) n \eta}{1+n(\eta-1)}\right)^{\eta/2}.
\end{equation}
The two results (\ref{eq:theta-star})-(\ref{eq:speedstar}) can be used in concert to improve optimization with reasonable speed, taking into account their explicit dependence on the parameters.   Note that the speed increases {\it exponentially} with $F_{min}-F_0$ as we increase $\eta$ while ${\bf \Theta}_*^2$ grows only linearly.  This strongly motivates including $F_0$ to generate speedier evolution without paying a significant price in proximity to ${\bf \Theta}=0$.

In the special case with $F_{min}-F_0$ strictly zero, the peak in measure is at  ${\bf \Theta}_*^2=0$ for all $\eta$; in that case, the $\eta$-dependence enters into how strongly peaked the measure is there (\ref{eq:measure-RLeta}).
In that case a clear limiting factor is that the speed near the minimum (specifically the region ${\bf\Theta}^2 F_2 <1$) dies exponentially with $\eta$.  This last effect is avoided in the generic case with nonzero $F_{min}-F_0$.    

In this generic case with $F_{min}-F_0$, at large dimension $n$ we observe a strong dependence on $\eta-1$ in  (\ref{eq:theta-star})-(\ref{eq:speedstar}).  The special choice $\eta=1$ leads to strong growth with $n$ of ${\bf \Theta}_*$ and  $|\dot{{\bf\Theta}}|_{{\bf\Theta}_*}$.  Conversely, choosing $\eta >1$ removes this behavior and increasing $\eta$ pushes the peak ${\bf \Theta}_*$ toward the optimum while maintaining speed $|\dot{{\bf\Theta}}|_{{\bf\Theta}_*}$.   
Thus we predict that $\eta>1$ should improve optimization performance per se, and we will test and confirm this below.  Of course, in ML problems the optimization performance (training loss) need not reflect the test accuracy, particularly in overparameterized problems, and we will analyze the effect of our hyperparameters (HP)s, including $\eta$, separately on training and test performance.  
In less over-parameterized ML settings, with $F_{min}>0$, this analysis indicates that $\eta>1$ helps directly with test accuracy as this reflects the optimization itself.  

We will verify aspects of these predictions in our experiments below and use it to improve ECD optimization in practice.
A limitation will be that we will set $F_0=0$, a simple way to  keep the number of tested HPs comparable among the different optimizers.

That is we will not experimentally exploit the role of $F_0$ in combination with $\eta$ just derived, in concentrating the measure while maintaining speed (\ref{eq:speedstar}), but we will observe a window where $\eta$ improves optimization performance in itself.  The joint behavior of $\eta, F_0$ along with the learning rate and chaos elements presents a promising avenue for further improvements.     Relatedly, we can improve the theoretical analysis by doing this section's calculations directly in the discrete case, e.g. summing rather than integrating over momenta in (\ref{eq:measure-general}).     

\section{The ECDSep Optimization Algorithm}
Here we construct our algorithm from a simple symplectic 1st order integration of the evolution equations from the Hamiltonian (\ref{eq:RLeta}), adding chaos-inducing elements and addressing mini batches.

\subsection{The update rules}\label{sec:upRules}

We can separate the Hamiltonian (\ref{eq:RLeta}) by taking a logarithm \cite{ESH}, obtaining
\begin{equation}
    H_{sep}(\mathbf{\Pi},\mathbf{\Theta}) = \log(\Pi^2)+\log(V[F(\mathbf{\Theta})])\,.
\end{equation}
Here $V$ is any function of the target $F$, and in this work we focus on a power law (\ref{eq:F-F0-eta}).
The dynamics induced in parameter space can explore any value of the objective $F$ for any $E$, by virtue of $\Pi$'s full range, with $|\Pi|\to 0$ for $F\to \infty$.
If it is known that the relevant part of the problem does not require exploring $F\to\infty$, the Hamiltonian can be regularized such that for a given $E$ only an upper-limited range of $F$ is explored.  This is achieved by a simple change of the kinetic term 
\begin{equation}
    H_{reg}(\mathbf{\Pi},\mathbf{\Theta}) = \log(\Pi^2+1)+\log(V[F(\mathbf{\Theta})])\,.
\end{equation}
Calling $\mathcal{E} \equiv e^E$, where $E$ is the energy associated to $H_{sep}$, the dynamics will now only explore regions in $\mathbf{\Theta}$ with $F(\mathbf{\Theta}) -F_0\leqslant \mathcal{E}^{1/\eta}$. The advantage is that the algorithm now has a smaller region to explore, but it will not be able to overcome arbitrarily high barriers in the target. This regularization does not change the dynamics in the accessed regions where $V\ll \mathcal{E}$, since $\Pi^2 \gg 1$. We will consider both algorithms, by adding a discrete switch $s = \{0,1\}$ that for $s = 1$ selects the regularized Hamiltonian. The continuum evolution equations are
\begin{equation}\label{eq:eoms}
    \dot{\Pi}_i = -\eta \frac{\partial_i F}{F-F_0}\:,\qquad \qquad \dot{\Theta}_i = \frac{2 \Pi_i}{s+\Pi^2}\,,
\end{equation}
where $\eta$ is the hyperparameter that concentrates the measure towards $F=F_0$ as in (\ref{eq:measureeta}). We consider a simple first order discretization that leads to the iterative update rules
\begin{equation}\label{eq:upRules}
\mathbf{\Pi}_{t+1}= \mathbf{\Pi}_t - \frac{\Delta t\, \eta}{F({\bf{\Theta}}_t)-F_0} \bm{\nabla}F({\bf{\Theta}}_t) \:,\qquad \mathbf{\Theta}_{t+1} = \mathbf{\Theta}_{t} +2 \Delta t \frac{\mathbf{\Pi}_{t+1}}{s+\Pi^2_{t+1}}\,,
\end{equation}
where $\Delta t$ is the step-size hyperaparameter. (Being the Hamiltonian separable, an interesting future direction would be to explore the use of 2nd order schemes that do not require extra gradient evaluations.) It is instructive to compare this to Gradient Descent with Momentum (GDM), see e.g. \cite{goh2017why}, which reads 
\begin{equation}
    \mathbf{\Pi}_{t+1} = \mathbf{\Pi}_{t}-(1-\beta)\mathbf{\Pi}_{t}-\bm{\nabla}F({\bf{\Theta}}_t)\,,\qquad \qquad \mathbf{\Theta}_{t+1} = \mathbf{\Theta}_{t}+\alpha \mathbf{\Pi}_{t+1} \,,
\end{equation}
where $\beta\in (0,1)$  is usually called the \emph{momentum} parameter (not to be confused with $\mathbf{\Pi}$) and $\alpha$ the learning rate. We write $\mathbf{\Pi}$'s update in this way to stress that for GDM $(1-\beta)$ acts as friction (a negative additive term in the update for the velocity, proportional to the velocity). Friction is how GDM converges: setting $\beta = 1$ removes the friction and ruins convergence.

For ECD instead convergence arises via non-linearities in the update rules: in the example \eqref{eq:RLeta} the evolution slows down as the mass increases. Without the need for friction to converge, energy is conserved; $\mathcal{E} = (F-F_0)^\eta (\Pi^2+s)$  is constant. More precisely $\mathcal{E}$ is exactly constant for the continuum dynamics (\ref{eq:eoms}) but in the discrete case it can oscillate about ${\cal E}$ by an amount controlled by $\Delta t$ (see \S\ref{sub:consEn}). Energy conservation   implies (\ref{eq:upRules}) $ \mathbf{\Theta}_{t+1} = \mathbf{\Theta}_{t} + \frac{2 \Delta t (F-F_0)^\eta}{\mathcal{E}}\mathbf{\Pi}_{t+1}$. Comparing with with GDM, we see that the dynamics behaves effectively as having a non-constant learning rate that adaptively changes according to the local value of the objective function $F$.
This is true also for $\mathbf{\Pi}$ (\ref{eq:upRules}): the magnitude of the gradient update is informed by the local value of the function.  In \S\ref{sec:experiments} thanks to this property the algorithm generically performs well without needing any HP scheduling. 

This is a general feature of ECD: it provides gradient based optimization with updates  informed by $F({\bf \Theta})$.
This leads to the HP $F_0$ introduced above in (\ref{eq:F-F0-eta}), which can be adaptively tuned as in \cite{BBI} and in the code associated to this work. One way to view this parameter is as the expected value of the target at the global minimum.  More precisely, its effect on the distribution of results and the (continuum) speed of propagation was derived above in (\ref{eq:measure-RLeta})-(\ref{eq:speedstar}).   
In our experiments below, $F_0 =0$, but a full HP analysis based on the formulae in \S\ref{sec:theory-formulaic} is an important future direction.

Finally, to follow the updated rules (\ref{eq:upRules}) we need to specify an initialization for the momenta $\mathbf{\Pi}_0$. Generically, we can initialize them along the direction of minus the initial gradient, with magnitude $|\mathbf{\Pi_0}|^2 \equiv \delta E$. With this choice, the energy is $\mathcal{E} = F_{init}^\eta(s+\delta E)$. For the regularized algorithm $s = 1$ we can set $\delta E=0$ (default) and it explores regions with $F\leqslant F_{init}$. A higher value of $\delta E>0$ allows the regularized algorithm to explore regions in parameter space where $F > F_{init}$.   For $s=0$,  we default to $\delta E = 1$. 
We initialize the momenta as
$    \mathbf{\Pi}_0 = \sqrt{\delta E} \frac{\bm{\nabla}F({\bf{\Theta}}_0)}{|\bm{\nabla}F({\bf{\Theta}}_0)|}$.

\subsection{Chaos-inducing elements and the volume formula}\label{sub:bounces}
The discrete dynamics in (\ref{eq:upRules}) is symplectic, meaning that volumes in phase space are \emph{exactly} conserved by the discrete dynamics (modulo numerical errors). See \cite[\S2]{leimkuhler2015molecular} for an introduction. Thanks to this, the continuum prediction (\ref{eq:measure-general}) for the distribution of results holds to a very good approximation also for the discrete dynamics, provided sufficient chaos. 

Although chaos is generic, it can fail or take too long to set in. 
We encourage chaotic exploration by a tunable modification ensuring trajectories disperse. In \cite{BBI}, energy-conserving billiard bounces were introduced with this purpose, implemented by randomly rotating the momentum after a certain number of steps (either fixed or dynamically tuned).
In this work, we simplify the chaos-inducing prescription (reducing the number of HPs) by adapting a method introduced in \cite{Robnik:2022bzs}:  we replace the full random momentum rotation after multiple steps with a random rotation by some angle at every step.
More precisely, after each step the momentum is further updated as
\begin{equation}
    \mathbf{\Pi} \leftarrow \frac{|\Pi| }{| \frac{\mathbf{\Pi}}{|\Pi|}+\nu \mathbf{z}|}\left(\frac{\mathbf{\Pi}}{|\Pi|}+\nu \mathbf{z}\right)
\end{equation}
where $\nu$ is the \emph{chaos hyperparameter} and each component of $\mathbf{z}$ is drawn from a standard gaussian. The prefactor ensures that the the norm of $\Pi$ is conserved, thus conserving $E$. 
As a consequence of this, the momentum will rotate by an amount controlled by $\nu \sqrt{n}$, where $n$ is the dimensionality of the problem \cite{Robnik:2022bzs}. When $\nu\sqrt{n}\ll 1$ the angle between the 
 original and rotated ${\bf \Pi}$ is small, and 
 for $\nu\sqrt{n} > 1$ the angle is $\pi/2$.  See  \S\ref{sub:nu-n-relations} for more details.

\subsection{Energy conservation with batches}\label{sub:consEn}
Though phase space volumes are exactly conserved by symplectic integration (\ref{eq:upRules}), energy violations $\sim \Delta t^2$ arise, just oscillating about the continuum value (until much later timescales with $E$ drift).

This applies to problems in which the target $F$ is fixed (i.e. full batch). When minibatches are present, they introduce explicit time dependence in the target,  $F(\bf{\Theta}(t);t)$ coming from the fact that at given time intervals the target explicitly changes since it is being evaluated on different batches of data.
As a result, $E$ will not be automatically conserved with minibatches. To enforce this, we rescale the $\mathbf{\Pi}$ at each step projecting back to the original energy surface, if possible. Explicitly after the update of $\mathbf{\Pi}$ we compute $\Pi^2$, and compare with the value it should have in the original energy surface, from the relation $\Pi^2 = \frac{\mathcal{\varepsilon}}{(F-F_0)^\eta}-s$, where we recall that $\mathcal{E}$ is the fixed energy determined at initialization. If the right hand side is positive we rescale the $\Pi$ homogeneously so that $\Pi^2$ agrees with it. 
We observe good performance from this on ML problems with minibatches. It can be skipped for optimization problems in which the target does not change in time, including full-batch training of neural networks.  

\subsection{Weight decay (WD)}
For a non-linear optimizer, WD and $L^2$ regularization act differently. WD is defined as exponential decay of the weights during training \cite{Hanson1988ComparingBF}, while the latter is the addition of a $\Theta^2$ term to the loss. While for SGD there is no difference between the two (with an appropriate rescaling) the difference appears already for Adam, as stressed in \cite{adamW} with regard to AdamW. 
More precisely, the first way of implementing WD changes the update rule of the parameter as
$    \mathbf{\Theta}_{t+1} = \dots  -\Delta t w_{d0}  \mathbf{\Theta}_t   $
where the dots denote terms that would be there in the update rule if weight decay was zero. This modification of the Hamiltonian dynamics introduces an $E$ violation.
The second one, akin to an $L^2$ term, is just a modification of the target function, not a change of the form of the Hamiltonian. To implement it, we define a constant $w_d$ and modify the update rules by the shift
\begin{equation}\label{eq:wd2}
    F(\mathbf{\Theta}) \to F(\mathbf{\Theta})+ \frac{w_d}{2} \Theta^2\,,\qquad\qquad {\bm\nabla} F(\mathbf{\Theta}) \to {\bm \nabla} F(\mathbf{\Theta})+w_d \mathbf{\Theta}\,.
\end{equation}
While the first choice can also be made compatible with the ECD framework when $E$ conservation is explicitly enforced  as in \S\ref{sub:consEn}, at this stage we have only implemented the $E$-conserving option in (\ref{eq:wd2}). It would be interesting to compare the two for ECD algorithms to see if it yields an advantage.

\subsection{The full algorithm}\label{sec:algorithm}

Algorithm \ref{alg:ECDSep} summarizes ECDSep, aside from the also-included adaptive tuning of $F_0$. We note here that {\bf Theorem 2.1} of \cite{BBI} -- the impossibility of stopping at a local minimum due to the energy conservation -- holds also in our current algorithm (see \S\ref{app:theoryAmdProofs} in the Appendix for our proof). 
The $s = 1$  algorithm explores the region in which the objective function $F$ satisfies $F-F_0 \leqslant (F_{init}-F_0)(1+\delta E)^{1/\eta}$, while the $s =0$ algorithm explores the whole space.

\begin{algorithm}[htb!]
   \caption{\small \emph{ECDSep}. $s = 1$ (default) is the regularized version of the algorithm. Defaults are $F_0 = \delta E = w_d = 0$, $\Delta t = 0.4$, $\nu = 10^{-5}$ and $\eta$ is required.  For $s = 0$, $\delta E = 1$.  $\varepsilon_1 = 10^{-10}$ and $\varepsilon_2 = 10^{-40}$ are numerical constants ensuring stability.  The block ensuring energy conservation can be removed for optimization problems without minibatches. }
   \label{alg:ECDSep}
\begin{algorithmic}
   \REQUIRE $F(\mathbf{\Theta})$: Function to minimize.
   
   \REQUIRE $\mathbf{\Theta}$: Initial parameter vector.
    \STATE $\mathcal{E} \gets \left(F(\mathbf{\Theta}) - F_0 + \frac12 w_d \mathbf{\Theta}^2\right)^\eta(\delta E+s)$ (Initialize energy)
    \STATE $\mathbf{\Pi} \gets - \frac{\bm{\nabla} F (\mathbf{\Theta})}{|\bm{\nabla} F (\mathbf{\Theta})|} \sqrt{\delta E}$ (Initialize momenta)
\REPEAT
    \STATE $V\gets \left(F(\mathbf{\Theta}) - F_0 + \frac12 w_d \mathbf{\Theta}^2\right)^\eta$ 
    
    \IF { energy conservation = True }
        \STATE $\pi_C^2 \gets \frac{\mathcal{E}}{V}-s$ 

    \IF{ $|\mathbf{\Pi}^2-\pi_C^2| > \varepsilon_1$ \AND $\pi_C^2 > 0$ }
    \STATE $\mathbf{\Pi} \gets \sqrt{\frac{\pi_C  ^2}{\mathbf{\Pi}^2}} \mathbf{\Pi}$ 
    \ENDIF
    \ENDIF
    \STATE $\mathbf{\Pi} \gets \mathbf{\Pi}- \frac{\Delta t \eta}{V^{1/\eta}} \left(\bm{\nabla} F(\mathbf{\Theta})+w_d \mathbf{\Theta}\right)$
    \STATE $\mathbf{\Theta} \gets \mathbf{\Theta}+2 \Delta t  \frac{\mathbf{\Pi}}{\mathbf{\Pi}^2+s}$
    \STATE $\mathbf{\Pi}_N \gets \frac{\mathbf{\Pi}}{|\mathbf{\Pi}|} + \nu \mathbf{z} $ \;($\mathbf{z}$ is a normal random vector)
    \STATE $\mathbf{\Pi}\gets \frac{|\mathbf{\Pi}|}{|\mathbf{\Pi_N|}} \mathbf{\Pi}_N$
 \UNTIL{$V< \varepsilon_2 $}
\end{algorithmic}
\end{algorithm}

\subsection{A guide to hyperparameter tuning}\label{sub:guideTuning}

Important HPs for the algorithm \ref{alg:ECDSep} include the step-size $\Delta t$, the concentration exponent $\eta$, and the chaos HP $\nu$; we expect also the loss offset $F_0$ to play an important role ultimately (cf. \S\ref{sec:theory-formulaic}). Along with the continuuum formulas in \S\ref{sec:theory-formulaic} which relate the concentration of measure to some of HPs, their intuitive behavior explained earlier in this section simplifies the task of their tuning.   In our experiments we will focus on $\Delta t, \eta,$ and $\nu$, leaving a detailed experimental treatment of $F_0$ for future work (while including an adaptive tuning option for $F_0$ in our current algorithm).   

As described in (\ref{eq:measureeta}), increasing $\eta$ from $1$ accentuates the region of smallest loss. We expect this to be important for problems in the undeparameterized regime, where the loss is a good indicator of the test accuracy, and thus where we want to get to the bottom of the basin as quickly as possible. We see indications of this in the problems in \S\ref{sub:graphs} and on the comparison in Fig.~\ref{fig:CIFAR100}.  The formulas in \S\ref{sec:theory-formulaic} suggest opportunities for significant principled improvements bringing in HP relations including $F_0$.

On the other hand, a  choice $\eta = 1$ can work for situations where instead of directly targeting the loss we want the optimizer to explore the low-loss basin region, with the goal of visiting points that might have a higher test accuracy (either individually or after averaging) even if they have a higher loss. We check this on image classification problems in \S\ref{sub:CIFAR}, where we show that weight averaging \cite{izmailov2018averaging} with $\eta = 1$ efficiently explores the low-lying basin finding solutions that generalize well, without needing any learning rate scheduling. 
More generally one could use our understanding of the HPs' effects on the measure to sample at the radius $|{\bf \Theta}|_*$ in (\ref{eq:theta-star}) to see how that affects performance.

Increasing $\nu$ accentuates chaos, generically reducing the mixing time. 
In a given problem it may need to to exceed a certain threshold so that (\ref{eq:measure-general}) applies.
As discussed above in \S\ref{sub:bounces}, the bounce angle per step is determined by $\nu^2 n$, with small-angle bounces for $\nu^2n \ll 1$ and large ($\pi/2$ angle) bounces otherwise.  
We find examples of each regime in our HP-tuned experiments below (see \S\ref{supp:hpscaling}), with the latter somewhat akin to a random walk.
 We set $10^{-5}$ as default, and did a logarithmic scan to understand it for different problems.  This again is an opportunity for improvement by relating it to the timescale related to $|{\bf \Theta}|_*$ in (\ref{eq:theta-star}), similarly to the typical set timescale giving a tuning-free algorithm in \cite{Robnik:2022bzs}.  We derive the corresonding relations in \S\ref{sub:nu-n-relations}.

We tune the learning rate $\Delta t$,  empirically finding good performance with a high value, setting $0.4$ as default. This effect might be related to the more stable symplectic (and approximately energy-conserving) dynamics. 
It will be an important future direction to derive a principled scaling of the hyperparameters with the scale of the problem, c.f. \cite{roberts2021principles,2022arXiv220303466Y}.

\section{Experimental results}\label{sec:experiments}

Here we compare ECDSep (Algorithm \ref{alg:ECDSep}) to SGD with momentum, Adam and AdamW, often used to achieve state of the art performance different ML tasks. We treat synthetic benchmarks in \S\ref{sec: synthetic_benchmark}, image classification  in \S\ref{sub:CIFAR}, node classification problems on graphs in \ref{sub:graphs} and natural language processing (NLP) in \S\ref{sub:BERT} .

The problems in \ref{sec: synthetic_benchmark} (and Fig.~\ref{fig:CIFAR100}) test ECDSep as a pure optimizer (targeting directly the loss), while the problems in the other sections analyze its ability to find regions in the loss landscape that lead to better accuracy in ML setups (e.g. overparameterized) where the two don't agree.  Along the way, we test (for just $F_0=0$) the
predicted improvement of the training loss as a function of $\eta$ in \S\ref{sec:theory-formulaic}.

For Adam, AdamW and SGD we scanned over a fixed number of parameters: the learning rate $\alpha$, WD and the momentum $\beta$ (only for SGD). Depending on the experiment, references can supply a good range of HPs for some optimizers; we then allow a larger scan for the others. We briefly discuss the setups and HP selection; more details appear in \S\ref{app:exp}, and the full code to reproduce them can be found at \url{https://github.com/gbdl/ECDSep}.
Given the modest statistics, we  only quote the average in the tables and present the full distribution of the corresponding experimental results in \S\ref{app:exp} in the Appendix.

The experiments in \S\ref{sec: synthetic_benchmark} were performed on a single laptop CPU, the experiments in \S\ref{sub:CIFAR} and \S\ref{sub:BERT} on a single NVIDIA-GEFORCE-RTX2080Ti with 11GB of GPU memory and 16GB RAM, and the experiments in \S\ref{sub:graphs}
on a single NVIDIA Tesla T4 with 15GB of GPU memory and 12.7GB of RAM. 

{{\bf Overall} we find that ECDSep, while not always best on individual experiments, is more reliable across a heterogeneous set of problems. Depending on the problem Adam(W) performs better than SGDM, or viceversa, with ECDSep competitive with the leader, as summarized in Table \ref{summary-table}. }
\begin{table}[!ht]
  \caption{\small Average accuracy over the experiments in this paper (excluding synthetics) \vspace{\baselineskip}}
  \label{summary-table}
  \centering
  \begin{tabular}{lcccc}
    \toprule
         & ECDSep & SGD & Adam & AdamW    \\
    \midrule
    Average accuracy &  \textbf{74.83} & 73.10 & 74.06 & 73.74 \\
    \bottomrule
  \end{tabular}
\end{table}
\subsection{Synthetic benchmarks}\label{sec: synthetic_benchmark}
\begin{figure}[h]
    \centering
    \subfloat{\includegraphics[width=0.5\textwidth]{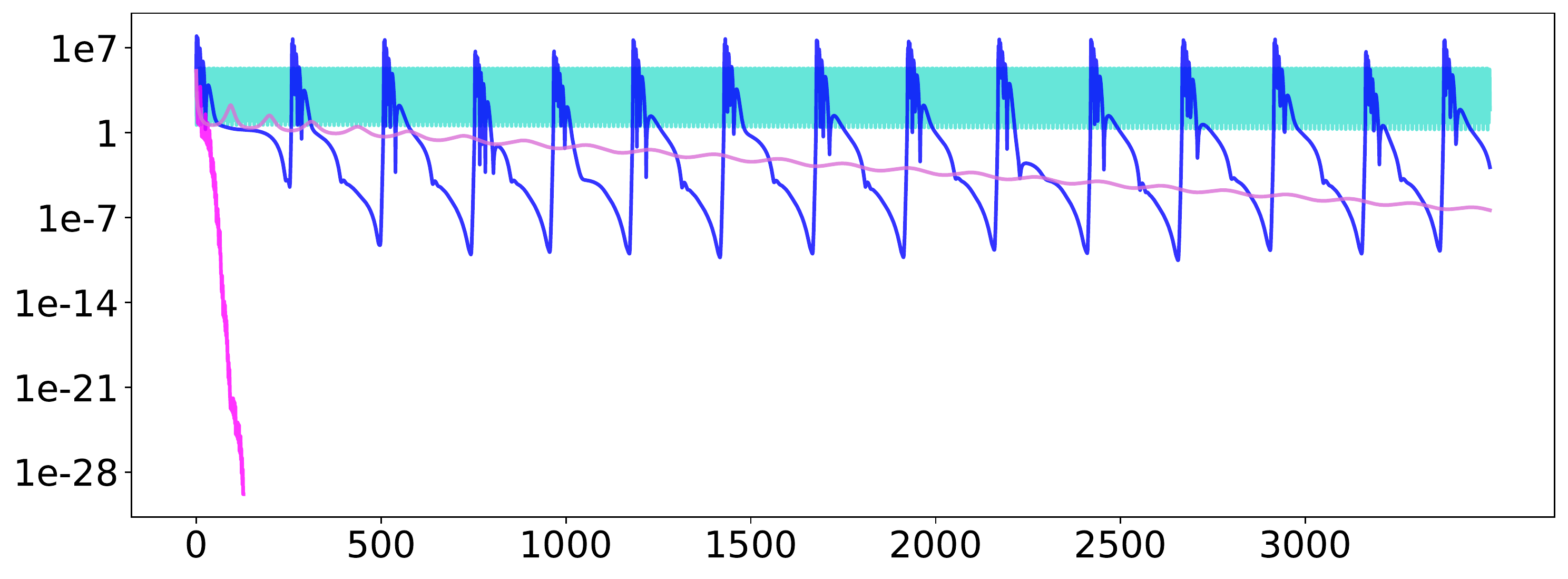}}%
    \subfloat{\includegraphics[width=0.5\textwidth]{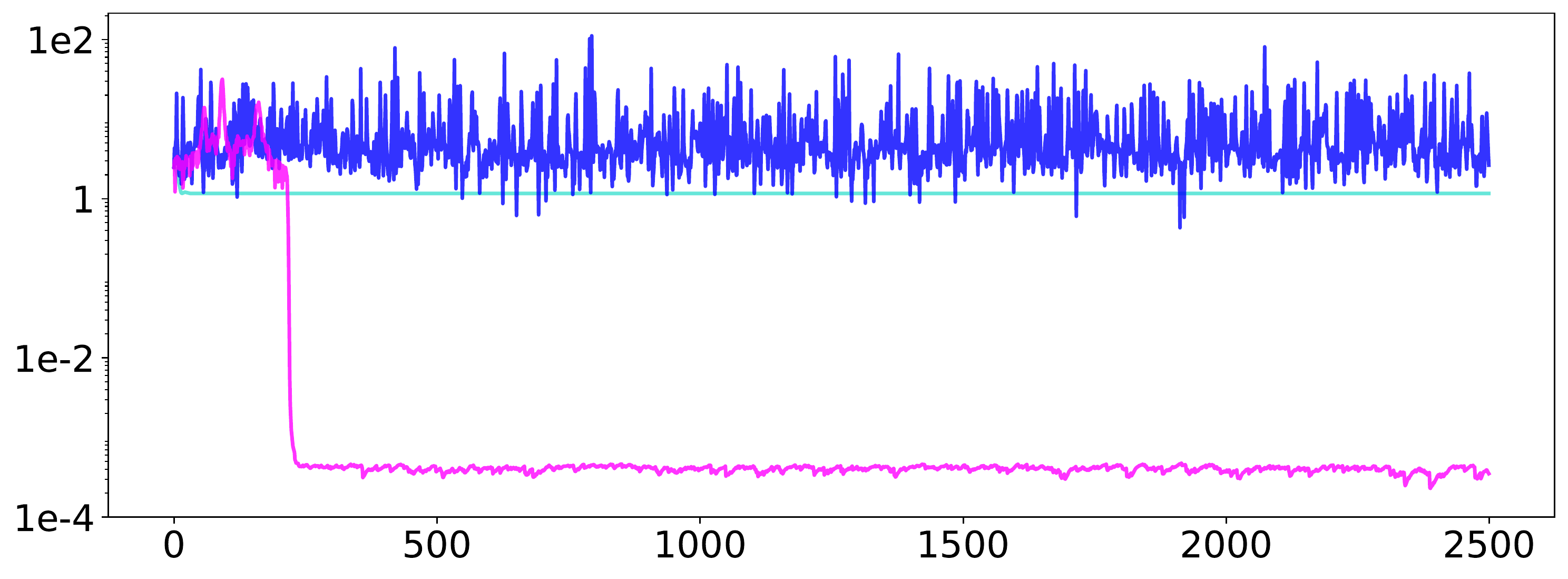}}\\
    \vspace{-.3cm}
    \subfloat{\includegraphics[width=0.5\textwidth]{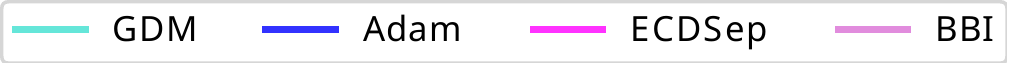}}
    \caption{\small Zakharov and Ackley: a typical run from a random starting point. }
    \label{fig:synth}
\end{figure}

We start by testing Algorithm \ref{alg:ECDSep} on synthetic benchmark optimization problems and compare with GDM and Adam (=AdamW since WD=0).  
To test on a problem with a shallow valley we considered the $n=10$ Zakharov function \cite[Function 173]{jamil2013literature}, where we also compared to BBI \cite{BBI}, and to test escape from local minima we use a regularized Ackley function \cite{ackley}. We select HPs using Optuna \cite{optuna_2019} by performing 500 trials starting form a fixed point, and
compared algorithms starting from new random initial points for more iterations (Fig.~\ref{fig:synth}). More details can be found in \S\ref{supp:synth}.

\subsection{Image classification}\label{sub:CIFAR}
\paragraph{CIFAR-100 and Tiny Imagenet training with weight averaging}

We trained residual networks from scratch on the popular medium-scale image classification tasks CIFAR-100 \cite{CIFAR} and Tiny-Imagenet \cite{tinyin}. 
State of the art on these problems with residual networks and without data augmentation \cite{izmailov2018averaging} uses Stochastic Weight Averaging (SWA), averaging the network visited during the last phase of the training, giving a network with a better test accuracy than the individual elements of the average.

Our goal is to understand how ECDSep compares with SGD and Adam(W) in this task. Since in this problem we are not interested in concentrating the volume at small loss but on exploring the low-loss region to find points with better accuracy, for this comparison we set $\eta = 1$(default) as explained in \S\ref{sub:guideTuning}; later we will analyze the training loss, to test the behavior of $\eta$ in optimization per se.

For CIFAR-100 we reproduced and improved the results in \cite[\S4.4]{izmailov2018averaging} by training a WideResnet28x10 \cite{Zagoruyko2016WideRN} with SGDM on CIFAR-100 and scanning on their same learning rates $\alpha=\{ 0.1, 0.05, 0.01, 0.001\}$  to which we added a momentum scan $\beta = \{0.9, 0.95, 0.99\}$.  From the procedure in that paper, we also borrowed the value of weight decay ($w_d = 5\times10^{-4}$), training epochs (300) and the fixed epoch at which to start the averaging,  \texttt{swa-start} = 161 (budget 1.5). More details on the SWA procedure are collected in \S\ref{supp:CIFAR}. For Adam and AdamW, we scanned over a broader set of learning rates $\alpha=\{ 0.1, 0.01, 0.001, 0.00001\}$ and allowed a possibly different weight decay $w_d = \{10^{-4}, 5\times10^{-4}, 10^{-3}\}$. For ECDSep instead, we scanned over learning rates $\Delta t = \{0.4, 0.6\}$ and chaos parameter $\nu = \{ 10^{-4}, 5\times 10^{-5}, 10^{-5}\}$. We repeated each experiment 4 times with different seeds, with best results in Table \ref{CIFAR100-table}.

To confirm these results, we performed a more extensive study on a different data set and with a different architecture: ResNet-18 \cite{Resnet} on Tiny Imagenet.
We trained for 100 epochs and, since we were using a constant learning rate thorough, we gave more freedom to each optimizer by allowing it to choose the best epoch at which the averaging starts. This is done by saving the networks found during training and averaging over them at the end, going back up to half of the training. The best result found is then kept. For SGD we performed an extensive scan on $\alpha = \{ 0.1, 0.05, 0.01, 0.001\}$ and $\beta = \{ 0.9, 0.99\}$ and fixed  $w_d = 10^{-4}$. For ECDSep we fixed the same value for the weight decay and fixed again $\eta = 1$ and scanned over $\Delta t = \{0.4, 0.6\}$ and chaos parameter $\nu = \{ 10^{-4}, 5\times 10^{-5}, 10^{-5}\}$. For Adam and AdamW we scanned  $\alpha = \{ 0.1, 0.01, 0.001, 0.0001\}$ and also allowed different weight decays values $w_d = \{ 10^{-4}, 5\times 10^{-4}\}$.  We repeated all the experiments three times with different seeds and collected the results in Table \ref{CIFAR100-table}.

\begin{figure}[h!]
    \centering
    \subfloat{{\includegraphics[width=\textwidth]{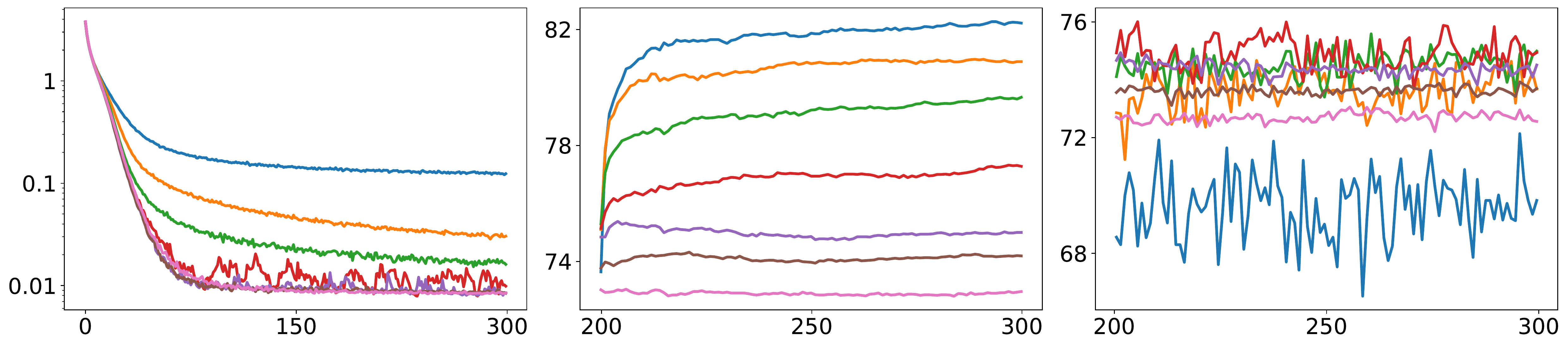} }}\\
    \vspace{-.3cm}\subfloat{{\includegraphics[width=\textwidth]{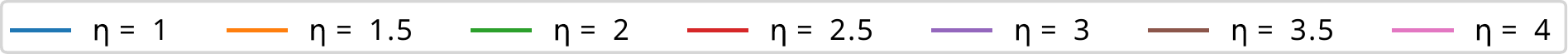} }}
    \caption{\small Left: training loss, center: SWA test accuracy, right: test accuracy, over 300 epochs.}
    \label{fig:CIFAR100}
\end{figure}

\begin{table}[!ht]
\caption{\small Full training on CIFAR-100 and TinyImagenet and fine-tuning on IN-1K: mean accuracy.
\vspace{\baselineskip}}
  \label{CIFAR100-table}
  \centering
  \begin{tabular}{lcccc}
    \toprule
         & ECDSep & SGD & Adam & AdamW\\
    \midrule
    CIFAR 100 & \textbf{82.57  } & 82.50 & 79.01 & 78.71\\
    Tiny Imagenet & \textbf{66.44 } & 64.83 & 61.67 & 59.84\\
    IN-1K (fine tuning) & 70.49 & 70.49 & 70.48 & 70.48\\
    \bottomrule
  \end{tabular}
\end{table}
\vspace{\baselineskip}
Finally, we analyzed the behavior of ECDSep as a function of $\eta$. In Fig.~\ref{fig:CIFAR100} we show different training runs with the WideResnet on CIFAR 100 for different values of $\eta$, keeping fixed the other hyperparameters ($\Delta t = 0.5$, $\nu = 10^{-4}$). We see that increasing $\eta >1$ monotonically decreases the loss as predicted in \S\ref{sec:theory-formulaic}, in this case tested up to $\eta = 4$. In this overparametrizied regime this translates into an improvement on accuracy only up to a certain point $\eta = 2.5$, with extra improvements on the loss being detrimental for accuracy. However, the $\eta = 1$ case, which has a lower accuracy before averaging, makes a much higher jump in accuracy after SWA (starting here at epoch 200), resulting in the best averaged accuracy. This suggests that the $\eta = 1$ dynamics visits a region of the landscape more relevant for generalization in this class of problems.

\paragraph{Fine tuning on Imagenet-1K}

We tested ECDSep on the task of fine-tuning on a large image dataset, Imagenet-1K \cite{deng2009imagenet}, which is freely available for non-commercial research and educational purposes. Due to computational constraints, we did not perform the whole tuning, but we start with a pre-trained ResNet 18 in PyTorch, which we fine-tuned for 10 extra epochs and perform SWA as in \S\ref{sub:CIFAR}.
We fix $w_d = 10^{-4}$ and again $\eta = 1$ and we scan $\Delta t = \{0.4, 0.1, 0.05, 0.01\}$ and $\nu =\{ 10^{-3}, 10^{-4}, 10^{-5}\}$. For SGD we scanned $\alpha = \{ 5\times 10^{-2}, 10^{-2}, 10^{-3}, 10^{-5} \}$ and momentum $\beta = \{0.9, 0.95,0.99\}$
For Adam and AdamW we allowed the weight decay to change, and we scanned over $\alpha = \{ 10^{-3}, 10^{-4}, 5\times10^{-5},10^{-5}\}$ and $w_d = \{10^{-3},10^{-4}, 10^{-5}\}$. Table \ref{CIFAR100-table} averages 2 runs per optimizer, which are essentially tied.

\subsection{Graphs}\label{sub:graphs}
We tested our optimizer on two graph problems from the Open Graph Benchmark (OGB) \cite{ogb}, a rich graph benchmark licensed under the MIT license. The datasets are \texttt{ogbn-arxiv} and \texttt{ogbn-proteins}, and in both cases a deep neural network is trained for node classification. In particular, we focus on certain Graph Neural Networks (GNNs) \cite{gdl_book,gnns_kipfWelling}. The details about the datasets and the GNNs are described in \S\ref{supp:graphs}.
For SGD and Adam(W) we performed a scan over $\alpha =\{10^{-1}, 5\cdot 10^{-2}, 10^{-2}, 5\cdot 10^{-3}, 10^{-3}, 5\cdot 10^{-4}\}$ and momentum for SGD over $\beta = \{0.9, 0.95, 0.99, 0.999\}$. Regarding ECDSep, first we did a few runs to understand the scale of the HPs, then we performed a finer scan over $\Delta t=\{1.5, 1.8, 2, 2.5, 2.8, 3\}$, $\eta=\{4.5, 5, 5.5, 6\}$, with $\nu=10^{-5}$; we scanned WDs $\{0, 10^{-3}, 10^{-4}, 10^{-5}\}$ for all 4. We performed $10$ and $5$ runs of the best combination of HPs with different seeds for \texttt{ogbn-arxiv}  and \texttt{ogbn-proteins} respectively. See Table \ref{tab:graph_table}. AdamW performed best, with ECDSep very close. For \texttt{ogbn-proteins} ECDSep outperforms SGD by many points of ROC-AUC score.

\begin{table}[!ht]
    \caption{\small Full training \texttt{ogbn-arxiv},  \texttt{ogbn-proteins}: best accuracy, ROC-AUC score. \vspace{\baselineskip}}
    \label{tab:graph_table}
  \centering
  \begin{tabular}{lcccc}
    \toprule
         & ECDSep & SGD & Adam & AdamW \\
    \midrule
    \texttt{ogbn-arxiv} &  71.55 &  71.81 & 72.37 &$\textbf{72.41}$\\
     \texttt{ogbn-proteins}& 74.67 & 65.79 & 77.42 & \textbf{77.44} \\
    \bottomrule
  \end{tabular}
\end{table}

\subsection{Language on the BERT transformer}\label{sub:BERT}

Next we finetune BERT \cite{devlin2018bert} on the GLUE benchmark \cite{wang2018glue},  one of the standard benchmarks for NLP (licensed under the CC BY 4.0 license). The GLUE benchmark comprises $8$ different NLP tasks: CoLA (acceptability prediction), MNLI (natural language inference), MRPC (semantic-similarity scoring), QNLI (sentence pair classification), QQP (semantic-similarity scoring), RTE (natural language inference), SST-2 (sentiment classification), STS-B (text scoring).

The scan over different optimizers was performed as follows. For Adam and AdamW we scanned the learning rate over the values identified in \cite[App.~A.3]{devlin2018bert} $\{ 2 \times 10^{-5}, 3 \times 10^{-5}, 5 \times 10^{-5}\}$, and the weight decay over $\{0, 10^{-2}, 10^{-3}\}$, giving 9 combinations in total. For SGD, since we had no prior information of performance on this problem, we allowed a more extensive scan on learning rates $\{10^{-2},10^{-3},10^{-4},10^{-5}\}$, momentum $\{0.9, 0.99\}$ and weight decay $\{10^{-2}, 10^{-3}\}$, for 18 combinations in total.
For ECDSep first, similarly as in the graph setting,  we briefly determined that a smaller learning rate was needed being this a fine tuning problem, fixing it at $\Delta t = 0.04$, then we scanned over $\eta = \{1,1.4, 2.0\}$,  $\nu  = \{10^{-4},10^{-5}\}$  and $w_d = \{0,10^{-2}\}$, for 12 combinations in total. For SGD, Adam and AdamW we used a linear learning rate schedule as in \cite[App.~A.3]{devlin2018bert}.
We repeated each experiment 3 times with results in Table \ref{BERT-table}, and in full in \S\ref{supp:BERT}.

\begin{table}[!ht]
  \caption{\small BERT fine tuning. First $8$ columns:  the average over $3$ runs of the metrics test Matthews correlation for CoLA, test Spearmans's correlation for STS-B, test F1 score for MRPC and QQP, and test accuracy for the remaining datasets. Last column: average over all datasets. \vspace{\baselineskip}}
  \label{BERT-table}
  \centering
  \begin{tabular}{lcccccccc|c}
    \toprule
         & MNLI     & QQP & QNLI & SST-2 & CoLA & STS-B & MRPC & RTE & avg. \\
    \midrule
    ECDSep & 84.24  & 86.70 & 91.19 &92.66 & 57.91 &89.26 & 90.96&73.16 & 83.26\\
    SGD &  83.31 & 86.36 & 91.03 & 92.17 & 60.54& 89.26 & 90.88&71.96 & 83.19 \\
    
    Adam & 84.31  & 88.14 & 91.39 &92.81 & 59.34 &89.02 & 91.09&71.36 & 83.43\\
    AdamW & 84.41  & 88.21 & 91.49 &93.03 & 59.68 &89.15 & 91.13&71.24 & \textbf{83.54}\\
    \bottomrule
  \end{tabular}
\end{table}
\vspace{\baselineskip}
\section{Limitations and Future Directions}

We described important limitations in the text, and here summarize.  We have a broad class of experiments, but low statistics and dominantly over-parameterized.  We have not exploited the full power of the predictions in \S\ref{sec:theory-formulaic} for the distribution of results and speed of propagation, particularly as regards the dependence on and relationships among $\eta, F_0, \nu$ and other HPs, and their dependence on the dimension $n$ (including width scalings). All this leaves room for substantial improvements in the future.  Those calculations could be generalized to the discrete case, though our symplectic integration controls some discretization errors.    Finally, we mention that adaptive methods to integrate the equations of motion \eqref{eq:eoms} can provide further improvements to performance, via the wealth of literature about numerical Hamiltonian integration. It would also be interesting to explore the possibility of algorithmically discover Hamiltonian ECD optimizers, along the lines of \cite{Chen2023SymbolicDO}.

\section{CO$2$ emission}
Overall, the experiments in \S\ref{sec:experiments} run for a cumulative of approximatively 11658 hours (of which 7672 hours for the experiments presented in the paper). Total emissions are estimated to be 683.34 kgCO$_2$eq, of which 17.14 kgCO$_2$eq directly offset. More details can be found in \S\ref{supp:CO2}.
\section{Acknowledgments}

We are grateful to Daniel Kunin for discussions and collaboration during the early stages of this project. We also thank Guy Gur-Ari, Ethan Dyer, Aitor Lewkowycz, Dan Roberts, Jakob Robnik, Uros Seljak, Sho Yaida and participants
of the 2023 Aspen Winter conference on ``Theoretical Physics for Machine Learning''  for fruitful discussions and suggestions. 
Our research is supported in part by the Simons Foundation Investigator and Modern Inflationary Cosmology programs, and the National Science Foundation under grant number PHY-1720397. 
Some of the computing for this project was performed on the Sherlock cluster. We would like to thank Stanford University and the Stanford Research Computing Center for providing computational resources and support that contributed to these research results.
\appendix
\section{More details on the theory and proofs} \label{app:theoryAmdProofs}

Here we derive (\ref{eq:measure-RLeta}) from the first equality in (\ref{eq:measure-general}), plugging in the Hamiltonian (\ref{eq:RLeta}).  This gives 
\begin{align}
    \mu &= \int d^n\Theta d^n\Pi \delta(E-V \Pi^2) \\
    &= \int d\Omega_{n-1} \int d^n\Theta \int d|\Pi| |\Pi|^{n-1} \delta(E-V|\Pi|^2)\\
    &= \int d\Omega_{n-1}\int d^n\Theta \frac{|\Pi|^{n-1}}{2 V |\Pi| }|_{|\Pi|=\sqrt{E/V}} \\
    &= \frac{E^{(n-2)/2}\pi^{n/2}}{\Gamma(n/2)}\int d^n\Theta V^{-n/2}.
\end{align}
The first equality is the definition of the measure.  The second rewrites the momentum integral in terms of its angular and radial directions (the Hamiltonian depends only on the radial direction, its magnitude).  In the third equality we used the delta function to do the integral over the magnitude $|\Pi|$ of the momentum, via the general formula $\int dx \delta(f(x)) = 1/f'(x)|_{x_*}$ where $f(x_*)=0$.  The final line implements the substitution indicated in the previous line and also evaluates the angular integral (the area of the $(n-1)$-sphere).

\begin{theorem}\label{thm:no-stopping}
If $V\ne 0$ (and $V\ne E/s$) then $\dot{\bf\Theta}\ne 0$ in the continuum evolution, and ${\bf\Theta}(t+\Delta t)\ne {\bf\Theta}(t)$ for the discrete algorithm.
\end{theorem}
The theorem and proof is a direct generalization of Thm 2.1 in \cite{BBI}, with the replacement of the Born-Infeld Hamiltonian with (\ref{eq:RLeta}) and use of the  resulting update rules for our case (\ref{eq:upRules}).  In particular  in comparison to equation (4) of \cite{BBI}, we have constant energy
\begin{equation}
    E = \frac{\dot\Theta^2}{V} + sV = V\Pi^2 + sV
\end{equation}
where $s=0$ for the dynamics that can reach everywhere, and $s=1$ for the regularized version, as described in the main text.
A similar result holds for a wide range of ECD Hamiltonians.

\subsection{Theory of the chaos parameter $\nu$ and additional hyperparameter relations}\label{sub:nu-n-relations}

In \S\ref{sec:theory-formulaic} we used the energy-conserving feature of ECD to derive key relations such as (\ref{eq:theta-star}) and (\ref{eq:speedstar}). Following \cite{Robnik:2022bzs} we can further relate the parameter dimensionality $n$ and various hyperparameters to the chaos hyperparameter $\nu$ defined in \S\ref{sub:bounces}.  

First it is useful to note that the bounce angle is determined by the combination $\nu\sqrt{n}$, as in \cite{Robnik:2022bzs}.  This is because $\nu$ appears in combination with $z$ in the chaos update in \S\ref{sub:bounces}, and in its component-wise Gaussian distribution the expectation value of $z_i z_j$ is $n\delta_{ij}$.  In more detail, the bounce angle is determined by
\begin{equation}
    {\bf \Pi}\cdot {\bf \Pi'} = |{\bf \Pi}||{\bf \Pi'}|\cos(\alpha_\nu), 
\end{equation}
with ${\bf \Pi'}$ the bounce-updated momentum
\begin{equation}
    \mathbf{\Pi'} = \frac{|\Pi| }{| \frac{\mathbf{\Pi}}{|\Pi|}+\nu \mathbf{z}|}\left(\frac{\mathbf{\Pi}}{|\Pi|}+\nu \mathbf{z}\right).
\end{equation}
This gives
\begin{equation}
    \cos(\alpha_\nu) = \frac{1+\nu {\bf z}\cdot{\hat{\mathbf{\Pi}}}}{|{\hat{\mathbf{\Pi}}}+\nu\mathbf{z}|}= \frac{1+\nu {\bf z}\cdot{\hat{\mathbf{\Pi}}}}{\sqrt{1+2\nu {\hat{\mathbf{\Pi}}}\cdot\mathbf{z}+\nu^2\mathbf{z}^2}}
\end{equation}
where $\hat{\mathbf{\Pi}}$ denotes the unit vector in the direction of $\mathbf{\Pi}$.
To estimate the angle we take an expectation value of this in the standard Gaussian ensemble for each component of $\mathbf{z}$.   

For large $\nu^2 n$ this gives an expected $\cos(\alpha_\nu)$ near 0, meaning a large per-step bounce angle $\sim \pi/2$.  Below we will see that this regime arises in some of our fine-tuning experiments (ImageNet 1K and the BERT examples with $\nu = 10^{-4}$).  

For $\nu^2 n\ll 1$, the angle is small, $\alpha_\nu \sim \nu\sqrt{n}$.  This corresponds to a small bounce per step, a regime which arises in our other experiments.

In the use of energy-conserving chaotic Hamiltonian dynamics for sampling, a tuning-free prescription for $\nu$ was derived in \cite{Robnik:2022bzs}.  This followed from the observation that the needed chaos should amount to one bounce per orbit on the typical set, in that context at a distance $\theta_{typ}\propto \sqrt{n}$.  If we adopt this idea in our case, replacing the typical set by our scale $|\mathbf{\Theta}_*|$, 
we have a timescale between full ($2\pi$) bounces
\begin{equation}
    \Delta t_{sep}=E\Delta t = E \frac{2\pi |\mathbf{\Theta}_*|}{|\dot{\mathbf{\Theta}}_*|}.
\end{equation}
In the small-angle case, this means an angle per step
\begin{equation}
    \alpha_\nu \sim 2\pi \frac{\Delta t}{\Delta t_{sep}}
\end{equation}
with $\Delta t$ our step size.  
Plugging in the explicit forms for these quantities given in \S\ref{sec:theory-formulaic}, for small angle and for $F_0=0, s=1$ we find for $n\gg 1$
\begin{equation}\label{eq:tune1}
    \nu_* \sqrt{n} \approx \Delta t \sqrt{\frac{F_2}{F_{min}}}\left(\frac{F_{min}}{F_{init}}\right)^{\eta/2} \left(\frac{\eta}{\eta-1}\right)^{\eta/2} (\eta-1)^{1/2}
\end{equation}
with $F_{init}$ the value of the objective at initialization. 
For $\eta = 1$ and $F_0\neq 0$, this suggests instead for $n \gg 1$:
\begin{equation}
    \nu_* \sqrt{n} \approx \Delta t \sqrt{\frac{F_2}{F_{init}-F_0}}\,.
\end{equation}

This analysis unveils strong dependencies among our hyperparameters.  For example, given the  small ratio $F_{min}/F_{init}$ we note an exponential sensitivity of $\nu_*$ to $\eta$ in \eqref{eq:tune1}.  As with the other general relations derived in \S\ref{sec:theory-formulaic}, it will be very interesting to test and exploit them experimentally.   

\subsection{Self-tuning of $F_0$}
Here we describe a version of Algorithm \ref{alg:ECDSep} where $F_0$ is automatically tuned, if the initial value is higher than the true global minimum. The idea is that in this situation there will be a step at which, due to discreteness, one iteration will try to jump to a negative $V$. This works only for the case in which $\eta$ is an odd integer, to which we now restrict ourselves. Here we present a linear shift of $F_0$, but another option can be an exponential backoff with some cutoff. The algorithm \ref{alg:ECDSep-self} as presented here is preliminary.

\begin{algorithm}[htb!]
   \caption{\small \small \emph{ECDSep-self tuning}. $s = 1$ (default) is the regularized version of the algorithm. Defaults are $F_0 = \delta E = w_d = 0$, $\Delta t = 0.4$, $\nu = 10^{-5}$.  For $s = 0$, $\delta E = 1$.  $\varepsilon_1 = 10^{-10}$ and $\varepsilon_2 = 10^{-40}$ are numerical constants ensuring stability.  The block ensuring energy conservation can be removed for optimization problems without minibatches.  $\eta$ (required) has to be an odd integer.}
   \label{alg:ECDSep-self}
\begin{algorithmic}
   \REQUIRE $F(\mathbf{\Theta})$: Function to minimize.
   
   \REQUIRE $\mathbf{\Theta}$: Initial parameter vector.
   \STATE $\Delta F_0 \gets 0$ (Initialize $F_0$ shift)
    \STATE $\mathcal{E} \gets \left(F(\mathbf{\Theta}) - F_0 + \frac12 w_d \mathbf{\Theta}^2\right)^\eta(\delta E+s)$ (Initialize energy)
    \STATE $\mathbf{\Pi} \gets - \frac{\bm{\nabla} F (\mathbf{\Theta})}{|\bm{\nabla} F (\mathbf{\Theta})|} \sqrt{\delta E}$ (Initialize momenta)
\WHILE{True}
    \STATE $V\gets \left(F(\mathbf{\Theta}) -( F_0+\Delta F_0) + \frac12 w_d \mathbf{\Theta}^2\right)^\eta$ 

    \IF {$V < \varepsilon_2$}
        \STATE $\Delta F_0 \gets  \Delta F_0+5 V$
    \ELSE
    \IF { energy conservation = True }
        \STATE $\pi_C^2 \gets \frac{\mathcal{E}}{V}-s$ 

    \IF{ $|\mathbf{\Pi}^2-\pi_C^2| > \varepsilon_1$ \AND $\pi_C^2 > 0$ }
    \STATE $\mathbf{\Pi} \gets \sqrt{\frac{\pi_C  ^2}{\mathbf{\Pi}^2}} \mathbf{\Pi}$ 
    \ENDIF
    \ENDIF
    
    \STATE $\mathbf{\Pi} \gets \mathbf{\Pi}- \frac{\Delta t \eta}{V^{1/\eta}} \left(\bm{\nabla} F(\mathbf{\Theta})+w_d \mathbf{\Theta}\right)$
    \STATE $\mathbf{\Theta} \gets \mathbf{\Theta}+2 \Delta t  \frac{\mathbf{\Pi}}{\mathbf{\Pi}^2+s}$
    \STATE $\mathbf{\Pi}_N \gets \frac{\mathbf{\Pi}}{|\mathbf{\Pi}|} + \nu \mathbf{z} $ \;($\mathbf{z}$ is a normal random vector)
    \STATE $\mathbf{\Pi}\gets \frac{|\mathbf{\Pi}|}{|\mathbf{\Pi_N|}} \mathbf{\Pi}_N$
    \ENDIF
    \ENDWHILE
\end{algorithmic}
\end{algorithm}

\section{More details on the experimental setup and results}\label{app:exp}
In this section, we expand on the results and setups for the experiments presented in the main text in \S\ref{sec:experiments}. In particular we discuss more details about the full specification of the problems together with the full distributions of the results whose average appear in the Tables in the main  text. We conclude in \S\ref{supp:CO2} with more details on the estimation of CO2 emissions. 

The full code needed to reproduce the experimental results, together with licensing specification, can be found at \url{https://github.com/gbdl/ECDSep}.

\subsection{Synthetic Experiments}\label{supp:synth}
\begin{figure}[h]
    \centering
    \subfloat{\includegraphics[width=0.45\textwidth]{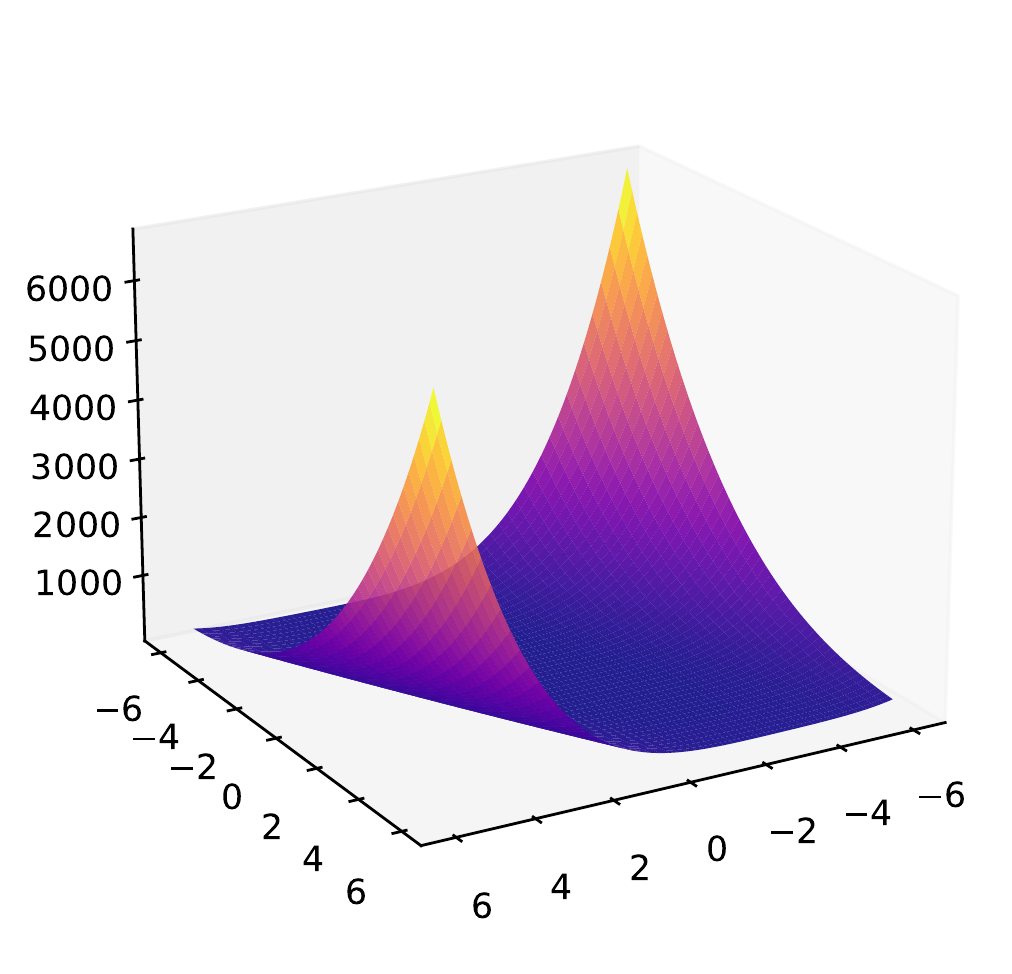}}%
    \subfloat{\includegraphics[width=0.45\textwidth]{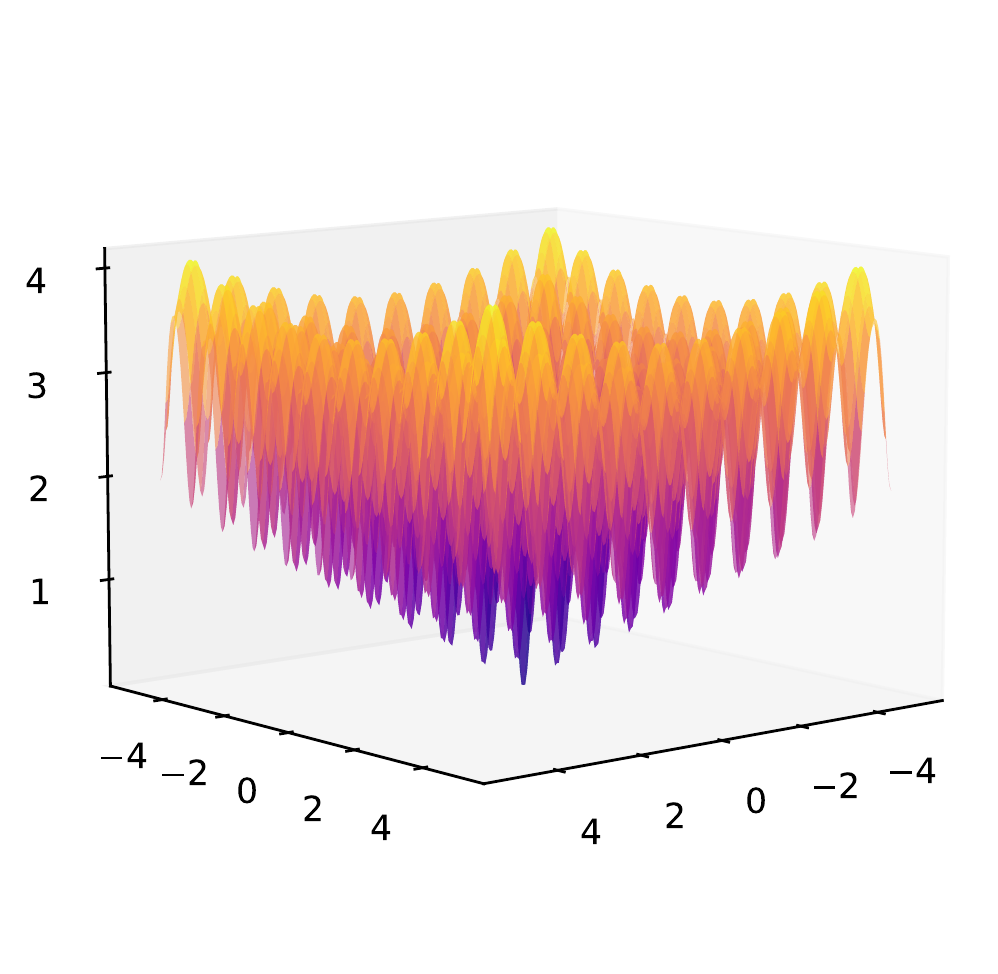}}\\
    \vspace{-.3cm}
    \caption{\small Left: a 2-dimensional depiction of the $n$-dimensional Zakharov function \eqref{eq:zakharov}; Right: regularized Ackley function \eqref{eq:ackley}.}
    \label{fig:ackley}
\end{figure}

The Zakharov function\cite[Function 173]{jamil2013literature} is a standard benchmark for optimization on shallow valleys, given by 
\begin{equation}\label{eq:zakharov}
    F(\mathbf{\Theta})\equiv\sum_{i=1}^n \theta_i^2+ \left(\frac12 \sum_{i=1}^n i \theta_i\right)^2+\left(\frac12 \sum_{i=1}^n i \theta_i\right)^4\;,
\end{equation}
and we study it for $n = 10$. It has no local minima, but the global minimum at $\mathbf{\Theta}=(0,\dots,0)$ lies in a nearly flat valley, slowing optimization. Even though we are studying it in 10 dimensions, to guide the eye we depict it for $n = 2$ in Fig.~\ref{fig:ackley}. To compare the various optimizers on such synthetic problems, we use Optuna\cite{optuna_2019} to find the best hyperparameters. 
Specifically we fix a common starting initial point $(1,\dots,1)$ and we perform 500 trials for each algorithm, each one corresponding to evolution for 250 iterations. With this approach, optimizers with more hyperparameters to tune are indirectly penalized since within the fixed amount of trials a much smaller portion of the search space is explored.

For Adam, we searched in $\alpha \in (10^{-2}, 10^{4})$, $\beta_1 \in(0.7, 1)$, $\beta_2 \in(0.7, 1)$, $\epsilon \in(10^{-12}, 10^{-6})$.
For ECDSep we searched in $\Delta t \in (10^{-2}, 10^{4})$, $\eta \in (1,4)$, $\nu \in (10^{-8},1)$, $\delta E \in (0,5)$ and consEn $\in \{\text{True},\text{False}\}$.
For SGD we first searched with learning rate in the same range of the other optimizer, but this always resulted in a divergent evolution. Thus we lowered the learning rate search space by searching over $\alpha  \in (10^{-8}, 10^{-3})$. The fact that SGD requires a smaller learning rate is confirmed by the fact that the optimal value found by Optuna for SGD is $\alpha \sim 10^{-6}$. Simultaneously we also searched over momentum $\beta \in (0.8, 1)$. Finally, for BBI we used the same search space as in \cite{BBI}. The results of a typical run from a randomly selected point using the HPs thus estimates are shown in Fig.~\ref{fig:synth}, left panel. More details can be found in the notebook \verb|ECDSep_zakharov.ipynb|. A $2$-dimensional depiction of the $n$-dimensional Zakharov function can be found in Fig. \ref{fig:ackley} (left). 

An important property of ECD algorithms is that thanks to energy conservation the evolution does not stop at local minima, as proved in Thm.~\ref{thm:no-stopping}. To check this and compare with common algorithms we tested evolution on a regularized two-dimensional Ackley function \cite{ackley}, which reads
\begin{multline} \label{eq:ackley}
    F(\theta_1, \theta_2) \equiv -20 \exp\left[-0.2\sqrt{0.5\left(\theta_1^{2}+\theta_2^{2}\right)}\right] 
 -\exp\left[0.5\left(\cos 2\pi \theta_1 + \cos 2\pi \theta_2 \right)\right]+\\ + e + 20 +10^{-8}(\theta_1^2+\theta_2^2)^4\,.
\end{multline}
We depict \eqref{eq:ackley} in Fig.~\eqref{fig:ackley} (right).

For this test we followed a protocol similar to the one used for the Zakharov function, this time starting from the initial point $(-4,3)$. For Adam, we searched over $\alpha \in (10^{-4}, 1)$, $\beta_1 \in(0.7, 1)$, $\beta_2 \in(0.7, 1)$, $\epsilon \in(10^{-12}, 10^{-6})$.
For ECDSep we searched in $\Delta t \in (10^{-4}, 1)$, $\eta \in (1,10)$, $\nu \in (10^{-5},1)$, $\delta E = 0$.
For SGD $\alpha  \in (10^{-8}, 10^{-3})$ and momentum $\beta \in (0.8, 1)$.

We find that SGD and Adam with a small learning rate immediately get stuck in local minima. With a higher learning rate they erratically explore the landscape, sometimes getting to smaller values of the target, but they do not converge there. ECDSep instead explores the landscape and eventually converges to the global minimum. The results of a typical run starting from a randomly selected point using the HPs thus estimates are shown in Fig.~\ref{fig:synth}, right panel. More details can be found in the notebook \verb|ECDSep_ackley.ipynb|.

\subsection{CIFAR-100 and Tiny Imagenet}\label{supp:CIFAR}

In the experiments in \S\ref{sub:CIFAR} we studied image classification using residual networks. In particular, we used the procedure of averaging the weights of the networks explored during the final phase of the training (SWA), which was introduced in \cite{izmailov2018averaging}, where it was demonstrated the the averaging results in a better test accuracy than the individual networks visited during training. See also \cite{guo2022stochastic} for a recent analysis and extension.

More in detail, \cite{izmailov2018averaging} discusses a procedure in which a first learning rate scheduling is performed for the pre-averaging phase, followed by the averaging phase that starts at a pre-determined epoch \texttt{swa-start} in which the learning rate is raised again and annealed. In \S\ref{sub:CIFAR}, we worked with a simplified protocol for SWA that requires much less tuning while still achieving very good performance, where the learning rate is kept constant during the whole training including the averaging phase. This has been studied already in \cite[\S4.4]{izmailov2018averaging} by training a WideResnet28x10 \cite{Zagoruyko2016WideRN} with SGD on CIFAR-100, where was shown that even though using a fixed high learning rate results in a much higher test error than with a scheduling, it makes a larger jump after averaging improving over SGD with fixed learning rate. We find the same phenomenon to be true for ECDSep where it is accentuated for small values of $\eta$, as shown in Fig.~\ref{fig:CIFAR100}.  The result we obtain for SGD is higher than the reported one (81.5) in  \cite[\S4.4]{izmailov2018averaging}, see Table \ref{CIFAR100-table}. As discussed in the main text, we also tested Adam and AdamW on the same dataset and with the same network.

For Tiny Imagenet, instead, we used a different network achitecture, ResNet 18 \cite{resnet-18}, and we did not fix beforehand the epoch at which the averaging starts, but we collected all the visited networks and at the end of training we computed running averaging start from the last epoch, for each experiment and for each optimizer. In each case we kept the best result found in this way. The motivation behind this protocol was to not to rely on the choice of \texttt{swa-start} as determined in \cite{izmailov2018averaging} where only SGD has been tested. The results are collected in Table \ref{CIFAR100-table} in the main text.

Below we present the best hyperparameters found with the scan discussed in \S\ref{sub:CIFAR}, together with the full distributions of results obtained by running the experiments multiple times with different seeds with the best HPs.
\begin{itemize}
    \item On CIFAR100:
    \begin{itemize}
        \item ECDSep: $\Delta t = 0.4, \nu = 5\times 10^{-5}$, with results $\{ 82.74, 82.6, 82.48, 82.45 \}$
        \item SGD: $\alpha = 0.05, \beta = 0.9$, with results $\{ 82.59, 82.59, 82.41, 82.39 \}$
        \item AdamW: $\alpha = 10^{-4}, w_d = 10^{-4}$, with results $\{ 79.04, 78.37, 78.66, 78.78\}$
        \item Adam: $\alpha = 10^{-4}, w_d = 10^{-4}$, with results $\{ 79.31, 79.1, 79.02, 78.58 \}$
\end{itemize}
    \item On Tiny Imagenet:
    \begin{itemize}
        \item ECDSep: $\Delta t = 0.6, \nu = 5\times 10^{-5}$, with results $\{ 66.42, 66.58, 66.33 \}$
        \item SGD: $\alpha = 0.1, \beta = 0.9$, with results $\{ 64.41, 64.64,64.45 \}$
        \item AdamW: $\alpha = 10^{-3}, w_d = 10^{-4}$, with results $\{ 59.84, 60.05, 60.12\}$
        \item Adam: $\alpha = 10^{-3}, w_d = 10^{-4}$, with results $\{ 61.59, 61.81, 61.6 \}$
    \end{itemize}
\end{itemize}

These experiments were conducted on a single NVIDIA-GEFORCE-RTX2080Ti with 11GB of GPU memory and 16GB RAM, with batch sizes 128.

\subsection{Imagenet-1K}\label{supp:IN}
For the experiments in this section, we performed a 10 epochs fine-tuning on the IN-1K dataset \cite{deng2009imagenet}, starting from a pre-trained ResNet-18 included with PyTorch. We then performed the same SWA procedure described for the Tiny-Imagenet experiments in \S\ref{supp:CIFAR}.

Due to resource constraints we only performed a short fine-tuning and with little statistics, which was insufficient to reveal appreciable differences among the optimizers. It would be interesting to perform a more extensive study of ECD on the IN-1K dataset.

For the scan detailed in \S\ref{sub:CIFAR}, we found the following best HPs and corresponding distribution of results.
\begin{itemize}
        \item ECDSep: $\Delta t = 0.1$ , $\nu  = 10^{-3}$, $w_d = 10^{-4}$, with results $\{ 70.515, 70.456 \}$
    \item SGD: $\alpha = 5\times 10^{-5}$, $\beta = 0.99$, $w_d=10^{-4}$, with results $\{ 70.494, 70.483 \}$
    \item AdamW: $\alpha =  10^{-5}$, $w_d = 10^{-4}$ with results, $\{ 70.481, 70.487 \}$
    \item Adam: $\alpha =  10^{-5}$, $w_d = 10^{-4}$ with results, $\{ 70.482, 70.479 \}$
\end{itemize}

These experiments were conducted on a single NVIDIA-GEFORCE-RTX2080Ti with 11GB of GPU memory and 16GB RAM, with batch sizes 128.

\subsection{Graphs}\label{supp:graphs}
The details about the datasets \texttt{ogbn-arxiv} and \texttt{ogbn-proteins} are given below.
\begin{itemize}
    \item \texttt{ogbn-arxiv} is a dataset made of a directed graph that represents the citation network between all Computer Science (CS) arXiv papers indexed by MAG \cite{mag}. Each vertex is an arXiv paper, and there is a directed edge between two vertices if one paper cites the other one. The goal is to predict the primary categories of arXiv CS papers, formulated as a $40$-class classification problem. The metric used to measure the performance is the usual accuracy. The graph has $169,343$ vertices and $1,166,243$ edges.
    \item \texttt{ogbn-proteins} is a dataset made of an undirected, weighted and typed graph. Each vertex is a protein and edges represent different types of biological association between proteins, like physical interaction, homology etc. The task is to predict the presence of a protein function among $112$ possibilities, so to label correctly each vertex. The metric used to mesure the performance is the ROC-AUC score. The graph has $132,534$ vertices and $39,561,252$ edges.  
\end{itemize}

For each of the above datasets the benchmark \cite{ogb} provides different types of deep neural networks. We focus on Graph Neural Networks (GNNs) having GraphSage \cite{graph_sage} as graph convolutional layers. Each GNN has $3$ graph convolutional layers with $256$ hidden channels and ReLU activation function. The GNN for \texttt{ogbn-arxiv} includes also batch normalization and $0.5$ droput. The number of training epochs is $500$ for \texttt{ogbn-arxiv} and $1000$ for $\texttt{ogbn-proteins}$, and the evaluation on the testing set is performed every $1$ and $5$ epochs respectively.

The best performance is
\begin{itemize}
\item for \texttt{ogbn-arxiv}: 
\begin{itemize}
    \item ECDSep: $\Delta t=2.8,\ \eta=4.5,\ \nu=10^{-5},\ w_d=0$, with results $\{71.57, 71.61, 71.9, \\ 71.62, 71.4, 71.43, 71.51, 71.53, 71.33, 71.48\}$.
    \item SGD: $\alpha=0.1$, $\beta=0.95$, $w_d=10^{-3}$ with results $\{71.64, 71.74, 71.51, 71.55, 71.37, \\ 71.58, 71.84, 71.49, 71.8, 71.39\}$.
    \item AdamW: $\alpha=5\cdot 10^{-3}$, $w_d=0$, with results $\{72.1, 72.65, 72.11, 72.42, 72.33, 72.34, \\ 72.38, 72.28, 72.42, 72.33\}$.
    \item Adam: $\alpha=5\cdot 10^{-3}$, $w_d=0$, with results $\{72.26, 72.11, 72.6, 71.95, 72.46, 72.41, \\ 72.42, 72.58, 72.54, 72.36 \}$.
\end{itemize}
\item for \texttt{ogbn-proteins}: 
\begin{itemize}
    \item ECDSep: $\Delta t=1.8,\ \eta=5,\ \nu=10^{-5},\ w_d=0$, with results $\{76.6, 74.61, 74.32,\\ 73.92, 73.88\}$.
    \item SGD: $\alpha=0.1$, $\beta=0.999$, $w_d=10^{-5}$, with results $\{68.32, 68.24, 66.1, 65.37, \\ 60.92\}$.
    \item AdamW: $\alpha=0.01$, $w_d=10^{-5}$, with results $\{78.09, 77.86, 77.81, 77.66, 77.53\}$.
    \item Adam: $\alpha=0.01$, $w_d=0$, with results $\{77.9, 77.87, 77.7, 77.52, 77.44\}$.
\end{itemize}
\end{itemize}

These experiments were performed on a single NVIDIA Tesla T4 with 15GB of GPU memory and 12.7GB of RAM. More details can be found in the notebooks \texttt{ECDSep\_graphs\_arxiv.ipynb} and \texttt{ECDSep\_graphs\_proteins.ipynb}.

\subsection{Language}\label{supp:BERT}

Here we report the HPs found with the scan discussed in \S\ref{sub:BERT} that resulted in the best performance. The metric used to determine the performance is different for different task, as discussed in Table \ref{BERT-table}. We also report here the full distribution of results over 3 initializations with different seeds. Recall that the learning rate $\Delta t = 0.04$ for all experiments with ECDSep.
\begin{itemize}
    \item For MNLI:
    \begin{itemize}
    \item ECDSep: $w_d = 0$, $\eta  = 2$ , $\nu  =  10^{-4}$, \{83.95, 84.38, 84.38\}
	\item SGD: $w_d = 10^{-3}$, $\alpha = 10^{-5}$, $\beta  = 0.99$, \{83.53, 83.52, 83.28\}
	\item AdamW: $w_d = 10^{-2}$, $\alpha = 2 \times10^{-5}$, \{84.56, 84.3, 84.36\}
	\item Adam: $w_d = 0$, $\alpha = 2\times10^{-5}$, \{84.16, 84.41, 84.36\}
\end{itemize}
\item For QQP:
\begin{itemize}
	\item ECDSep: $w_d = 0$, $\eta  = 2$ , $\nu  =  10^{-4}$, \{87.1, 86.53, 86.48\}
	\item SGD: $w_d = 10^{-3}$, $\alpha = 10^{-5}$, $\beta  = 0.99$, \{86.11, 86.02, 85.63\}
	\item AdamW: $w_d = 10^{-2}$, $\alpha = 2 \times10^{-5}$, \{88.31, 88.19, 88.12\}
	\item Adam: $w_d = 0$, $\alpha = 2 \times10^{-5}$, \{88.38, 88.21, 87.84\}
 \end{itemize}
\item For QNLI:
\begin{itemize}

	\item ECDSep: $w_d = 0$, $\eta  = 1.4$ , $\nu  =  10^{-5}$, \{91.16, 91.38, 91.03\}
	\item SGD: $w_d = 10^{-3}$, $\alpha = 10^{-5}$, $\beta  = 0.99$, \{91.25, 91.16, 90.77\}
	\item AdamW: $w_d = 10^{-2}$, $\alpha = 2 \times10^{-5}$, \{91.62, 91.43, 91.43\}
	\item Adam: $w_d = 0$, $\alpha = 2\times 10^{-5}$, \{91.49, 91.4, 91.29\}
\end{itemize}
\item For SST-2:
\begin{itemize}
	\item ECDSep: $w_d = 0$, $\eta  = 1$ , $\nu  =  10^{-4}$, \{93, 92.78, 92.2\}
	\item SGD: $w_d = 10^{-3}$, $\alpha = 10^{-5}$, $\beta  = 0.99$, \{92.43, 91.86, 91.86\}
	\item AdamW: $w_d = 0$, $\alpha = 2\times 10^{-5}$, \{93.23, 93, 92.89\}
	\item Adam: $w_d = 0$, $\alpha = 2 \times10^{-5}$, \{93.00, 92.78, 92.66\}
 \end{itemize}
\item For CoLA:
\begin{itemize}
	\item ECDSep: $w_d = 10^{-2}$, $\eta  = 2$ , $\nu  =  10^{-5}$, \{60.07, 57.85, 55.82\}
	\item SGD: $w_d = 10^{-3}$, $\alpha = 10^{-4}$, $\beta  = 0.9$, \{58.8, 58.11, 54.96\}
	\item AdamW: $w_d = 10^{-3}$, $\alpha = 3 \times10^{-5}$, \{59.87, 59.78, 59.38\}
	\item Adam: $w_d = 10^{-5}$, $\alpha = 2\times 10^{-5}$, \{61.09, 58.84, 58.08\}
 \end{itemize}
\item For STS-B:
\begin{itemize}
	\item ECDSep: $w_d = 10^{-2}$, $\eta  = 2$ , $\nu  =  10^{-5}$, \{89.3, 89.27, 89.21\}
	\item SGD: $w_d = 10^{-2}$, $\alpha = 10^{-5}$, $\beta  = 0.99$, \{88.72, 88.71, 88.68\}
	\item AdamW: $w_d = 0$, $\alpha = 3\times 10^{-5}$, \{89.37, 89.1, 88.98\}
	\item Adam : $w_d = 0$, $\alpha = 2 \times10^{-5}$. \{89.33, 89.09, 88.66\}
 \end{itemize}
\item For MRPC:
\begin{itemize}
	\item ECDSep: $w_d = 10^{-3}$, $\eta  = 1.4$ , $\nu  =  10^{-5}$, \{91.58, 90.18, 91.12\}
	\item SGD: $w_d = 10^{-3}$, $\alpha = 10^{-4}$, $\beta  = 0.99$, \{89.66, 88.19, 88.07\}
	\item AdamW: $w_d = 0$, $\alpha = 2\times 10^{-5}$, \{91.87, 91.07, 90.46\}
	\item Adam: $w_d = 10^{-2}$, $\alpha = 2\times 10^{-5}$, \{91.78, 90.78, 90.69\}
 \end{itemize}
\item For RTE:
\begin{itemize}
	\item ECDSep: $w_d = 10^{-3}$, $\eta  = 1$ , $\nu  =  10^{-5}$, \{75.09, 72.56, 71.84\}
	\item SGD: $w_d = 10^{-3}$, $\alpha = 10^{-4}$, $\beta  = 0.99$, \{71.12, 69.68, 69.31\}
	\item AdamW: $w_d = 10^{-2}$, $\alpha = 3\times 10^{-5}$, \{71.84, 71.12, 70.76\}
	\item Adam: $w_d = 10^{-2}$, $\alpha = 2 \times10^{-5}$, \{72.2, 71.84, 70.04\}
\end{itemize}
\end{itemize}

These experiments were conducted on a single NVIDIA-GEFORCE-RTX2080Ti with 11GB of GPU memory and 16GB RAM. More details can be found in the notebook \verb|ECDSep_language_bert|.
\subsection{Hyperparameter scaling}\label{supp:hpscaling}
An important goal for the new designed optimizer ECDSep is to understand how to scale the hyperparameters, in particular $\Delta t$, $\eta$ and $\nu$, with the problem size. In Table \ref{tab:hypers}, for each performed experiment we collect the values of the ECDSep HPs $\Delta t$, $\eta$ and $\nu$ together with the number of parameters of the associated neural network.  We stress that these are very different kind of problems, which include fine-tuning, weight averaging and small Graph Neural Networks. For these results we targeted test accuracy (which as discussed in the main text is not the same as optimization performance per se, which also 
favors larger $\eta$ e.g. in the image problems)

\begin{table}[h!]
\begin{center}
\caption{\small ECDSep HP values, number of parameters of the neural network and problem size for each experiment. For \texttt{bert} the HPs are averaged over all the GLUE datasets. (FT) after a dataset means that the goal was fine-tuning instead of training from scratch.\vspace{\baselineskip}}\label{tab:hypers}
\begin{tabular}{ccccc}
Experiment & Num. parameters & \multicolumn{3}{c}{Hyperparameters} \\ 
 & $n$ & $\Delta t$ & $\eta$ &  $\nu$ \\ \hline
\texttt{ogbn-arxiv} & $200$K  & $2.8$ & $4.5$ & $10^{-5}$\\
\texttt{ogbn-proteins} & $200$K & $1.8$ & $5$ & $10^{-5}$\\
\texttt{Tiny Imagenet} & $12$M & $0.6$ &$1$ & $5\times 10^{-5}$ \\
\texttt{Imagenet-1K} (FT) & $12$M &$0.1$& $1$& $10^{-3}$ \\
\texttt{CIFAR100} & $56$M &  $0.4$& $1$ & $5\times 10^{-5}$ \\
\texttt{bert} (FT) & $110$M & $0.04$ & $1.6$  & $4\times 10^{-5}$  \\
\end{tabular}
\end{center}
\end{table}

We note that $\Delta t$ decreases with bigger neural networks.  
It is also important to note the $\nu\sqrt{n}$ figure of merit determining the stepwise bounce angle as discussed in \S\ref{sub:nu-n-relations}.
Among the experiments here, we find some with small angle $\nu\sqrt{n}\ll 1$ and others among the fine-tuning ones (Imagenet-1K and some of the BERT examples, those with $\nu\sim 10^{-4}$) with $\nu\sqrt{n}\ge 1, \alpha_\nu\simeq \pi/2$ (a large bounce each step).  In these cases, such near-random-walk behavior proves competitive with traditional optimizers. 
This pattern deserves further study.

\subsection{CO2 Emissions}\label{supp:CO2}
Experiments in \S\ref{sub:CIFAR} and \S\ref{sub:BERT} used a private infrastructure with carbon efficiency $\sim 0.25$ kgCO$_2$eq/kWh with approximatively 10638 hrs computation (of which 7092 for the experiments in the paper) was performed on a GPU RTX 2080 Ti (TDP of 250W). Total emissions for these are estimated to be 666.2 kgCO$_2$eq.  Experiments in \S\ref{sub:graphs} used Google Cloud Platform in region us-west2, which has a carbon efficiency of 0.24 kgCO$_2$eq/kWh with approximatively 1020 hrs of computation  (of which 680 for the experiments in the paper) performed on a GPU T4 (TDP of 70W). Total emissions for these are estimated to be 17.14 kgCO$_2$eq of which 100\% were directly offset by the provider.  Estimations conducted using the impact calculator presented in \cite{lacoste2019quantifying}.

\bibliographystyle{plain} 
\bibliography{biblio}

\begin{thebibliography}{10}

\bibitem{ackley}
D.~Ackley.
\newblock {\em A Connectionist Machine for Genetic Hillclimbing}.
\newblock The Springer International Series in Engineering and Computer
  Science. Springer US, 2012.

\bibitem{optuna_2019}
Takuya Akiba, Shotaro Sano, Toshihiko Yanase, Takeru Ohta, and Masanori Koyama.
\newblock Optuna: A next-generation hyperparameter optimization framework.
\newblock In {\em Proceedings of the 25th {ACM} {SIGKDD} International
  Conference on Knowledge Discovery and Data Mining}, 2019.

\bibitem{2017arXiv170102434B}
Michael {Betancourt}.
\newblock {A Conceptual Introduction to Hamiltonian Monte Carlo}.
\newblock {\em arXiv e-prints}, page arXiv:1701.02434, January 2017.

\bibitem{gdl_book}
Michael~M. Bronstein, Joan Bruna, Taco Cohen, and Petar Veličković.
\newblock Geometric deep learning: Grids, groups, graphs, geodesics, and
  gauges, 2021.

\bibitem{Chen2023SymbolicDO}
Xiangning Chen, Chen Liang, Da~Huang, Esteban Real, Kaiyuan Wang, Yao Liu, Hieu
  Pham, Xuanyi Dong, Thang Luong, Cho-Jui Hsieh, Yifeng Lu, and Quoc~V. Le.
\newblock Symbolic discovery of optimization algorithms.
\newblock {\em ArXiv}, abs/2302.06675, 2023.

\bibitem{BBI}
Giuseppe~Bruno De~Luca and Eva Silverstein.
\newblock Born-infeld ({BI}) for {AI}: Energy-conserving descent ({ECD}) for
  optimization.
\newblock In Kamalika Chaudhuri, Stefanie Jegelka, Le~Song, Csaba Szepesvari,
  Gang Niu, and Sivan Sabato, editors, {\em Proceedings of the 39th
  International Conference on Machine Learning}, volume 162 of {\em Proceedings
  of Machine Learning Research}, pages 4918--4936. PMLR, 17--23 Jul 2022.

\bibitem{deng2009imagenet}
Jia Deng, Wei Dong, Richard Socher, Li-Jia Li, Kai Li, and Li~Fei-Fei.
\newblock Imagenet: A large-scale hierarchical image database.
\newblock In {\em 2009 IEEE conference on computer vision and pattern
  recognition}, pages 248--255. Ieee, 2009.

\bibitem{devlin2018bert}
Jacob Devlin, Ming-Wei Chang, Kenton Lee, and Kristina Toutanova.
\newblock Bert: Pre-training of deep bidirectional transformers for language
  understanding.
\newblock {\em arXiv preprint arXiv:1810.04805}, 2018.

\bibitem{goh2017why}
Gabriel Goh.
\newblock Why momentum really works.
\newblock {\em Distill}, 2017.

\bibitem{guo2022stochastic}
Hao Guo, Jiyong Jin, and Bin Liu.
\newblock Stochastic weight averaging revisited.
\newblock {\em arXiv e-prints}, pages arXiv--2201, 2022.

\bibitem{graph_sage}
William~L. Hamilton, Rex Ying, and Jure Leskovec.
\newblock Inductive representation learning on large graphs.
\newblock NIPS'17, page 1025–1035, Red Hook, NY, USA, 2017. Curran Associates
  Inc.

\bibitem{Hanson1988ComparingBF}
Stephen~Jose Hanson and Lorien~Y. Pratt.
\newblock Comparing biases for minimal network construction with
  back-propagation.
\newblock In {\em NIPS}, 1988.

\bibitem{Resnet}
Kaiming He, X.~Zhang, Shaoqing Ren, and Jian Sun.
\newblock Deep residual learning for image recognition.
\newblock {\em 2016 IEEE Conference on Computer Vision and Pattern Recognition
  (CVPR)}, pages 770--778, 2015.

\bibitem{resnet-18}
Kaiming He, Xiangyu Zhang, Shaoqing Ren, and Jian Sun.
\newblock Deep residual learning for image recognition.
\newblock In {\em Proceedings of the IEEE conference on computer vision and
  pattern recognition}, pages 770--778, 2016.

\bibitem{ogb}
Weihua Hu, Matthias Fey, Marinka Zitnik, Yuxiao Dong, Hongyu Ren, Bowen Liu,
  Michele Catasta, and Jure Leskovec.
\newblock Open graph benchmark: Datasets for machine learning on graphs, 2020.

\bibitem{izmailov2018averaging}
Pavel Izmailov, Dmitrii Podoprikhin, Timur Garipov, Dmitry Vetrov, and {Andrew
  Gordon} Wilson.
\newblock Averaging weights leads to wider optima and better generalization.
\newblock In Ricardo Silva, Amir Globerson, and Amir Globerson, editors, {\em
  34th Conference on Uncertainty in Artificial Intelligence 2018, UAI 2018},
  34th Conference on Uncertainty in Artificial Intelligence 2018, UAI 2018,
  pages 876--885. Association For Uncertainty in Artificial Intelligence
  (AUAI), 2018.

\bibitem{jamil2013literature}
Momin Jamil and Xin-She Yang.
\newblock A literature survey of benchmark functions for global optimisation
  problems.
\newblock {\em International Journal of Mathematical Modelling and Numerical
  Optimisation}, 4(2):150--194, 2013.

\bibitem{kingma2017adam}
Diederik~P. Kingma and Jimmy Ba.
\newblock Adam: A method for stochastic optimization.
\newblock {\em CoRR}, abs/1412.6980, 2015.

\bibitem{gnns_kipfWelling}
Thomas~N. Kipf and Max Welling.
\newblock Semi-supervised classification with graph convolutional networks,
  2017.

\bibitem{CIFAR}
Alex Krizhevsky.
\newblock Learning multiple layers of features from tiny images.
\newblock 2009.

\bibitem{lacoste2019quantifying}
Alexandre Lacoste, Alexandra Luccioni, Victor Schmidt, and Thomas Dandres.
\newblock Quantifying the carbon emissions of machine learning.
\newblock {\em arXiv preprint arXiv:1910.09700}, 2019.

\bibitem{tinyin}
Ya~Le and Xuan~S. Yang.
\newblock Tiny imagenet visual recognition challenge.
\newblock 2015.

\bibitem{leimkuhler2015molecular}
Ben Leimkuhler and Charles Matthews.
\newblock Molecular dynamics.
\newblock {\em Interdisciplinary applied mathematics}, 39:443, 2015.

\bibitem{adamW}
Ilya Loshchilov and Frank Hutter.
\newblock Decoupled weight decay regularization.
\newblock In {\em International Conference on Learning Representations}, 2018.

\bibitem{adamw_paper}
Ilya Loshchilov and Frank Hutter.
\newblock Decoupled weight decay regularization, 2019.

\bibitem{roberts2021principles}
Daniel~A. Roberts, Sho Yaida, and Boris Hanin.
\newblock {The Principles of Deep Learning Theory}.
\newblock 6 2021.

\bibitem{Robnik:2022bzs}
Jakob Robnik, G.~Bruno De~Luca, Eva Silverstein, and Uro\v{s} Seljak.
\newblock {Microcanonical Hamiltonian Monte Carlo}.
\newblock 12 2022.

\bibitem{ESH}
Greg~Ver Steeg and Aram Galstyan.
\newblock Hamiltonian dynamics with non-newtonian momentum for rapid sampling.
\newblock {\em CoRR}, abs/2111.02434, 2021.

\bibitem{sgd_paper}
Ilya Sutskever, James Martens, George Dahl, and Geoffrey Hinton.
\newblock On the importance of initialization and momentum in deep learning.
\newblock In {\em Proceedings of the 30th International Conference on
  International Conference on Machine Learning - Volume 28}, ICML'13, page
  III–1139–III–1147. JMLR.org, 2013.

\bibitem{wang2018glue}
Alex Wang, Amanpreet Singh, Julian Michael, Felix Hill, Omer Levy, and Samuel~R
  Bowman.
\newblock Glue: A multi-task benchmark and analysis platform for natural
  language understanding.
\newblock {\em arXiv preprint arXiv:1804.07461}, 2018.

\bibitem{mag}
Kuansan Wang, Zhihong Shen, Chiyuan Huang, Chieh-Han Wu, Yuxiao Dong, and
  Anshul Kanakia.
\newblock {Microsoft Academic Graph: When experts are not enough}.
\newblock {\em Quantitative Science Studies}, 1(1):396--413, 02 2020.

\bibitem{2022arXiv220303466Y}
Greg {Yang}, Edward~J. {Hu}, Igor {Babuschkin}, Szymon {Sidor}, Xiaodong {Liu},
  David {Farhi}, Nick {Ryder}, Jakub {Pachocki}, Weizhu {Chen}, and Jianfeng
  {Gao}.
\newblock {Tensor Programs V: Tuning Large Neural Networks via Zero-Shot
  Hyperparameter Transfer}.
\newblock {\em arXiv e-prints}, page arXiv:2203.03466, March 2022.

\bibitem{Zagoruyko2016WideRN}
Sergey Zagoruyko and Nikos Komodakis.
\newblock Wide residual networks.
\newblock {\em ArXiv}, abs/1605.07146, 2016.

\end{thebibliography}

\end{document}